\documentclass[review]{elsarticle}

\usepackage{xr}
\biboptions{sort&compress}
\usepackage{url}

\usepackage{hyperref}
\usepackage{mathrsfs}
\usepackage{epsfig}
\usepackage{epstopdf}
\usepackage[nosumlimits]{amsmath}
\usepackage{amssymb}
\usepackage{pifont}
\usepackage{amsfonts}
\usepackage{appendix}
\usepackage{subfigure}
\usepackage{longtable}
\usepackage{mathbbol}
\usepackage{algorithm}
\usepackage{caption}
\usepackage{color}
\usepackage{algpseudocode}
\usepackage{graphicx,subfigure}
\algrenewcommand\alglinenumber[1]{\small #1:}
\renewcommand{\a}{{\bf a}}
\renewcommand{\b}{{\bf b}}

\newcommand{\x}{{\bf x}}
\newcommand{\y}{{\bf y}}

\newcommand{\A}{{\bf A}}
\newcommand{\B}{{\bf B}}

\newcommand{\I}{{\bf I}}

\newcommand{\K}{{\bf K}}
\renewcommand{\L}{{\bf L}}

\newcommand{\X}{{\bf X}}
\newcommand{\Y}{{\bf Y}}

\newcommand{\btheta}{\boldsymbol{\theta}}

\newcommand{\bSigma}{\boldsymbol{\Sigma}}

{\end{list}}
\usepackage{amssymb,amsthm}
\usepackage{amsmath}
\usepackage{amsfonts}
\usepackage{breqn} 

\long\def\symbolfootnote[#1]#2{\begingroup%
\def\thefootnote{\fnsymbol{footnote}}\footnote[#1]{#2}\endgroup}

\newcommand{\be}{\begin{equation} }

\newcommand{\ee}{\end{equation} }

\newcommand{\undertilde}[1]{\mathord{\vtop{\ialign{##\crcr
    $\hfil\displaystyle{#1}\hfil$\crcr\noalign{\kern1.5pt\nointerlineskip}
    $\hfil\tilde{}\hfil$\crcr\noalign{\kern1.5pt}}}}}
\newfont{\ma}{msbm10}

\usepackage[numbers]{natbib}

\usepackage{color}

\begin{document}

\begin{frontmatter}


\title{Enhanced Gaussian Process Dynamical Models with Knowledge Transfer for Long-term Battery Degradation Forecasting}

\author[sheffield]{Xing, W.W.}
\author[szu]{Zhang, Z.}
\author[cqu]{Shah, A.A.$^{*}$}

\cortext[cor1]{Corresponding author: $^*${\it E-mail}:
akeelshah@cqu.edu.cn.}

\address[cqu]{Key Laboratory of Low-grade Energy Utilisation Technologies and Systems, MOE, Chongqing University, Chongqing 400030, China}

\address[sheffield]{School of Mathematics and Statistics, University of Sheffield, Hounsfield Road, Sheffield S3 7RH, United Kingdom}
\address[szu]{College of Computer Science and Software Engineering, Shenzhen University, Shenzhen, 518060, China}

\begin{abstract}

Predicting the end-of-life or remaining useful life of batteries in electric vehicles is a critical and challenging problem, predominantly approached in recent years using machine learning to predict the evolution of the state-of-health during repeated cycling. To improve the accuracy of predictive estimates, especially  early in the battery lifetime, a number of algorithms have incorporated features that are available from data collected by battery management systems. Unless multiple battery data sets are used for a direct prediction of the end-of-life, which is useful for ball-park estimates, such an approach is infeasible since the features are not known for future cycles. In this paper, we develop a highly-accurate method that can overcome this limitation, by using a modified  Gaussian process dynamical model (GPDM). We introduce a kernelised version of GPDM for a more expressive covariance structure between both the observable and latent coordinates. We combine the approach with transfer learning to track the future state-of-health up to end-of-life. The method can incorporate features as different physical observables, without requiring their values beyond the time up to which data is available. Transfer learning is used to improve learning of the hyperparameters using data from similar batteries.  The accuracy and superiority of the approach over modern benchmarks algorithms including a Gaussian process model and deep convolutional and recurrent networks is demonstrated on three data sets, particularly at the early stages of battery lifetime.
\\
\\
{\bf Word Count:} 6811
\vspace{0.25in}
\end{abstract}

\end{frontmatter}
\newpage

	\begin{tabular}{c | c c c c c}
		\hline
		 Symbol  & Meaning  \\
		\hline
		$\a,\b$     &  Vectors of basis expansion coefficients\\
		$\A,\B$     &  Matrices of basis expansion coefficients\\
		${\cal D}$    & Datasets\\
		$\K$     &  Column or row covariance matrix\\
		$k(\cdot,\cdot|\cdot)$     &  Kernel function\\
		$\L$     &  Cholesky decomposition factor\\
		${\cal L}$    & Likelihood function\\
		$m$     &  Battery label\\
				$n$     & Cycle number \\
				$\bf n$     & Normal noise \\
						$w$ & Weight factor for noise\\
		$\x$     &  Latent/hidden variable\\
		$\y$     &  Observable\\
		$\X$     &  Matrix of latent variables\\
		$\Y$     &  Matrix of observed data\\
		$\btheta$     & vector of hyperparameters\\
		$\theta$     & Hyperparameter\\
				$\pmb\Lambda$ &Gaussian process predictive variance\\
		$\pmb\mu$ &Gaussian process mean estimate of observable/latent variable\\
		$\sigma$ & Noise variance\\
		$\pmb\Sigma$     & Covariance matrix \\
		$\pmb \psi,\,\pmb\phi$     &  Basis functions\\
		\hline 
			\end{tabular}

\vspace{0.1in}			
	\begin{tabular}{c | c c c c c}
		\hline
		 Acronym  & Meaning  \\
		\hline
		CNN     &  Convolutional neural network\\
		DNN     &  Deep neural network\\
		EOL     & End of life \\
		GP     &  Gaussian process\\
		GPDM     &  Gaussian process dynamical model\\
		GPDM     & Gaussian process latent variable model \\
		LSTM     & Long short term memory \\
		RMSE     & Root mean square error \\
		RNN     &  Recurrent neural network\\
		RUL     & Remaining useful life\\
		SOH     & State of health \\
		\hline 
			\end{tabular}


\section{Introduction}
\setlength{\baselineskip}{24pt}

Batteries are essential components of modern energy systems, providing energy storage capabilities in a broad range of sectors.
The most prominent emerging application is electric vehicles, which are forecasted to become the primary method of transportation in the developed world beyond 2030. 
The main battery technologies for these applications are lithium-ion batteries (LiBs), which undergo a number of degradation processes during operation \citep{XIONG2020110048,LI2022119030}, leading to an irreversible loss in capacity and a need to replace the battery stack when the end-of-life (EOL) is reached. The EOL is normally defined as  the point at which the battery reaches 70-80 \% of its rated capacity, after which it can be used for second-life applications \cite{CHENG2023113053}.

One of the challenges in developing electric vehicles, therefore, is devising strategies to monitor, control and predict the future state of the battery stack \cite{RAUF2022111903}, as part of a broader attempt to develop digital twins \cite{NASERI2023113280}. This can aid maintenance, lower the risk of hazards such as fire and explosion, and inform the decision to retire the stack. At the heart of such strategies are algorithms that are able to predict the  state-of-health (SOH), a measure of the battery health condition, for future charge-discharge cycles \cite{LI2022119030,XIONG2020110048,tian2020review}. Such algorithms provide running predictions of the EOL and remaining useful life (RUL), namely the number of cycles remaining before the battery stack reaches EOL.

Algorithms for SOH prediction can be devised in a number of ways, with many of the early attempts reliant on physics-based or semi-empirical models of the degradation process,  and more recent models taking a purely  data-driven approach. Physics-based models are hampered by the fact that the  degradation phenomena (thermal, mechanical, electrochemical and chemical) are poorly understood.
Semi-empirical or mechanistic approaches generally use equivalent-circuit models that relate the circuit parameters to the SOH  via empirically-derived constitutive laws, with the  fitting of various parameter-dependent functions to data \cite{zhang2019real,HE201110314,SARASKETAZABALA2016839}.
The applicability and accuracy of these models is questionable since they rely on a reductive approach to modelling the degradation through functional forms and associated parameters, neither of which are likely to be transferrable. Moreover, the parameters are not necessarily constant as the degradation proceeds, and nor as operating conditions change, leading to the additional  requirement of tracking variations in the circuit parameters using data-driven methods such as a Kalman filter \cite{Bhangu2005783,HU2014182}.

Data-driven approaches for battery degradation forecasting have emerged as the preferred choice due to the increasing availability of battery cycling data as well as the aforementioned difficulties associated with physics-based and mechanistic approaches. The data is usually gathered from repeated charge-discharge cycling, predominantly  in constant-current  regimes. It takes the form of the capacity or SOH at a particular state-of-charge (SOC) as a function of cycle number. It may also include temperature, current, voltage and (ex-situ) impedance measurements.  The capacity is normally measured using Coulomb counting and the SOH is normally defined as the ratio of capacity to the rated or initial capacity.

 There are different types of data-driven approaches, which can be categorised in different ways. The simplest approach seeks to map the cycle number (input) to the SOH or capacity (output), after which the SOH at future cycles is predicted \cite{RICHARDSON2017209}. More sophisticated approaches augment the input (cycle number) with features related to the charge-discharge cycling performance \cite{YANG2018387,9558770}. A third approach uses data from multiple batteries, usually of the same type operating under the same load and operating conditions \cite{severson2019data}, with the RUL or EOL as the output.  A vast array of (supervised) machine learning methods have been employed, with the most common being Gaussian process (GP) models \cite{LIU2013832,RICHARDSON2017209,CHENG2023113053}, support vector regression (SVR) \cite{ZHAO201899,NI2022117922}, and  artificial neural networks (ANNs) or their deep equivalents (DNNs) \cite{MA2018102,WANG2020227591,LI2021230024}.  

In particular, DNNs have proven to be the most popular \cite{ZHANG2022112282}, with a  variety of architectures explored, including recurrent networks (RNNs) such as the long short term memory (LSTM)  or gated recurrent unit (GRU)  \cite{zhang2018long,LU2022,chemali2017long}, convolutional networks (CNNs) \cite{qian2021convolutional,DING2021111287}, transformer and encoder-decoder models based on RNNs \cite{9714323,LI2021230024},  and hybrid architectures such as a CNN-LSTM \cite{zraibi2021remaining} or CNN-GRU \cite{FAN2020101741}.  Although DNNs are unrivalled in (mostly unsupervised) pattern recognition tasks,  they require larger datasets compared to popular  non-parametric methods such as GP models~\cite{RICHARDSON2017209}, kernel regression and SVR, placing a greater demand on data acquisition and resources. The latter methods are also much simpler and faster to train by virtue of the much smaller number of model choices and parameters or hyperparameters.  Furthermore, there are few principled approaches towards architecture design and hyperparameter selection for DNNs, with the state-of-the-art being randomised or deterministic grid searches, feasibly  covering only a small portion of the entire space of possibilities for complex networks. 

Measurements of the charge-discharge voltage, current and temperature, along with impedance data can be used to extract features to feed as inputs into data-driven models. The  features are either hand crafted or learned using automated feature extraction methods. For instance,
Yang et al. \cite{YANG2018387} employed a GP model with hand-crafted physical features, including the duration of the charge phase and the slope of the voltage profile at the end of charge. Similarly, Chen et al. \cite{chen2021combining} used an RNN fed with up to 20 such physical features. 
Hong et al. \cite{HONG2020115646}, on the other hand, used a dilated CNN with sequences of the terminal voltage, current, and temperature as inputs to predict the RUL, extracting  features from the sequences via convolutional layers. Similarly,  Chemali \cite{chemali2017long} et al. used an LSTM with voltage, current, and temperature sequences to predict the SOH. 

The great drawback of feature engineering approaches for predicting the SOH is that the features are not available for future cycles, rendering these supervised machine learning approaches  valid only for a one-cycle-ahead prediction without using costly direct-embedding (multiple model) or sequence to-sequence approaches to learn very long sequences. For example, in Fan et al. \cite{FAN2020101741}, the charge voltage, temperature and current sequences at each cycle are used as features, making prediction of the SOH beyond $N+1$ infeasible, where $N$ is the number of cycles for which training data is available. Precisely the same issue arises in the model of Qian et al. \cite{qian2021convolutional}, which uses randomly selected segments of the charge voltage, differenced voltage and current as features.

Using multiple battery datasets  \cite{HONG2020115646,HSU2022118134}, as opposed to a single data set specific to a battery under investigation, can help to improve the transferability of a model to different conditions, and simultaneously allows for the use of features. Batteries, however, exhibit random variabilities in performance that can obviously not be predicted. The application of multiple battery data sets is therefore most appropriate when rough estimates of the EOL or RUL are required for a batch of batteries, as opposed to accurate online estimation of the SOH \cite{severson2019data,shu2020stage}. Thus, the majority of models using such data sets are focused on the EOL or RUL and do not attempt to predict the SOH or capacity trajectory. Moreover, most employ features from a specified number of early cycles. For example, Severson et al. used changes in the discharge voltage curves between specified cycles as predictors for the RUL.  Li et al. \cite{LI2021230024}, on the other hand, used a sequence-to-sequence model to make early-, mid- and late-life predictions of the EOL, with inputs defined by the capacity sequence up to a certain cycle.  This approach worked well for late predictions (the last 10\% or so of cycles) but was not very accurate for early predictions. 

Time-series methods, usually based on autoregression,  are in principle more appropriate for SOH prediction, being as it is a sequence problem.  Linear state-space  models such as the Kalman filter \cite{Bhangu2005783,duong2018heuristic} and ARIMA \cite{zhou2016lithium,KIM2022103888} have been used in this context, with varying degrees of success. The ARIMAX model of Kim et al. \cite{KIM2022103888} used an exogenous input that is not available for future cycles, restricting its use to single-cycle predictions. Khaleghi et al. \cite{KHALEGHI2021116159} similarly used randomly selected charge-voltage sequences to feed into a NARX model employing a multi-layer perceptron (MLP).

Nonlinear state space models that track a hidden state of the system are often employed for state-of-charge estimation, usually based on particle filters \cite{YE2018789} or extended  \cite{plett2004extended} and unscented  \cite{yu2017lithium}  Kalman filters. The use of such approaches for SOH estimation is rare, with a notable exception being the hidden Markov model (HMM) of Zhao et al.  \cite{ZHAO2022124632}, albeit for a satellite battery.  An alternative nonlinear state space model that is simpler to specify and train than HMMs is  the unsupervised Gaussian process dynamic model (GPDM) \cite{wang2005gaussian, wang2007gaussian}. GPDM is an extension of the Gaussian process latent variable model (GPLVM) \cite{lawrence2005probabilistic} that is often used for learning low-dimensional embeddings \cite{taylor2006modeling,reid2016cell}. The addition of Markov model for the hidden variable leads to the GPDM, which can be used to  predict sequential data. Amongst other applications, it has been used successfully for predicting human pose and motion \cite{urtasun20063d},  cloth manipulation \cite{koganti2017bayesian} and dynamic texture synthesis \cite{zhu2016dynamic}, and it is naturally able to handle missing data \cite{taylor2006modeling}.

In this paper, we develop a novel approach to address all of the issues described above. (i) the low accuracy of long-term forecasts (at least the last 50\% before EOL); (ii) the use of multiple data sets without compromising accuracy; (iii) incorporating features  for long-term SOH predictions; (iv) avoid the use of complex networks or DNNs that are difficult to design, have high model variance and place a larger burden on data acquisition. Our approach uses a modified  GPDM  that  eliminates the assumption of independence across coordinates of both the features and the latent variable. The resulting covariance matrix across basis functions can be integrated out and the model can be generalised using kernel substitution to a potentially infinite-dimensional approximating subspace (dependent on the choice of kernel).  The covariances across coordinates are then encoded using a Cholesky decomposition, leading to a richer covariance structure. 

We use the modified GPDM to map  the SOH and other quantities such as voltage and temperature of similar batteries  to a shared low-dimensional latent (hidden)  space.   
The features and latent variables evolve according to posterior GPs, yielding their trajectories up to an arbitrary horizon. In this way, future values of the features are no longer required. Furthermore, we use transfer learning to enhance the predictions, employing complete data sets (to EOL) for batteries from the same family and cycled under the same conditions to better predict the SOH for a specific battery (the target task). In contrast to other multi-dataset approaches, this allows for battery-specific data to be injected into the learning process.  

To demonstrate the early prediction capability of our method, we compare it to state-of-the-art GP and deep learning models enhanced by transfer learning on three  different  datasets. We employ strategies with and without physical features that are readily obtained from battery management systems. The results show that  our method is superior in terms of accuracy, especially for long term predictions. In conclusion, it can form the basis of online algorithms that can provide accurate predictions of RUL and EOL early in the battery life.

\section{Methods}

\subsection{Problem Definition}\label{probdef}

The discharge capacity of the battery $c(n)$ for a given cycle $n$ can be used as the basis for defining the state-of-health (SOH), namely $\mbox{SOH}(n)=c(n)/c(0)$. The EOL and RUL for the current cycle are defined as
\begin{equation}
\mbox{EOL}=\min n\;\;\;\mbox{such that} \;\;\mbox{SOH}(n)\le \vartheta, \quad \mbox{RUL}=\mbox{EOL}-n
\end{equation}
for some threshold $\vartheta$ normally in the range  $\vartheta\in(0.7,0.8)$.  Our primary goal is to predict the curve of $\mbox{SOH}(n)$ accurately and efficiently, especially early in the evolution, namely for cycles $n\le \frac{1}{2}\mbox{EOL}$.

Let $\y_{n} \in \mathbb{R}^{d}$ be a time-series comprising $d$ {\/\em attributes\/} of the battery at the $n$-th cycle. In the simplest case, $d=2$ and the only attributes are the SOH  and the cycle number, $\y_{n}=[n,\mbox{SOH}(n)]^\top$. In the more general case $d>2$,  $\y_{n}=[n,a_1(n),\hdots,a_{d-1}(n)]^\top$ with $a_1(n)=\mbox{SOH}(n)$ and additional scalar or vector attributes  $a_2(n),\hdots$, e.g., temperature and voltage values (or curves) from the charge or discharge cycles. In this case we seek to predict $a_1(n)$ for $n>T$   given training data ${\cal D}_T=\{\y_n\}_{n=1}^T$. In the development and evaluation of the model, we are provided  with a complete data set ${\cal D}_N=\{\y_n\}_{n=1}^N$, $N>T$, up to or beyond EOL. We shall consider only scalar attributes, although the method can be readily extended to the vector case.

Regarding the use of the term `features', which is the common terminology in the literature on battery degradation, we point out that in feature engineering, features are extracted from data, the components of which are referred to as attributes.  In particular, physical quantities that are used as inputs to battery degradation models (as opposed to features learned from these quantities)  will be referred to as attributes in this paper. The components of the latent variables in GPDM, on the other hand, are referred to as features, since they map the attributes in the data to a feature or latent space. 

A secondary goal in this study is the incorporation of attributes for multi-step ahead SOH prediction. When focussing on the SOH, in contrast to the RUL or EOL, incorporating attributes as inputs is infeasible since they are unknown for future cycles. In our model this will no longer present a problem for reasons that are explained in section \ref{GPDM}. A  third goal is to leverage data from other batteries of the same type and operated under the same conditions in order to predict  $\mbox{SOH}(n)$ more accurately for a given battery.  This is explained the next section. 

\subsection{Transfer learning}\label{probdef}

We follow the approach of Richardson et al. \cite{richardson2017gaussian}, in which the authors use data from multiple cells  by employing what they term a multi-output GP. More specifically, we are provided with data 
${\cal D}_{N_m}^m=\{\y_n^m\}_{n=1}^{N_m}$, $m=1,2,\hdots, M$, in which $N_m$ is the number of cycles for battery $m=1,2,\hdots, M$, and $\y_n^m=\mathbb{R}^{d}$ is the corresponding vector of attributes at cycle $n$. 
For simplicity, we set $N_m=N$ for all batteries, without loss of generality. The primary task is to predict $a_1^{m_*}(n)=\mbox{SOH}^{m_*}(n)$, the second attribute of  $\y_n^{m_*}$ for a given battery $m=m_*$, from cycles $T+1$ to EOL. For this task, we utilise the data
\begin{equation}
{\cal T}^{m_*}=\left(\bigcup_{m\neq m_*}{\cal D}_{N}^m\right)\cup {\cal D}_{T}^{m_*}
\end{equation}
That is, we use all of the data for $m\neq m_*$ and only the first $T$ cycles for $m=m_*$. 

To implement the above training procedure, we encode the outputs for all batteries as follows
\begin{equation}\label{encodingd}
\y_n=[n,m,a_1(n),\hdots,a_{d-1}(n)]^\top,
\end{equation}
so that the separate outputs $\y_n^m$ are replaced by a single output, in which the battery label $m$ is used to distinguish between battery-specific data. The GPDM method we propose is described in the next section.

\subsection{Enhanced Gaussian process dynamical models}\label{GPDM}
\begin{figure}[h]
	\centering
	\subfigure
	{\label{fig:c}\includegraphics[width=90mm]{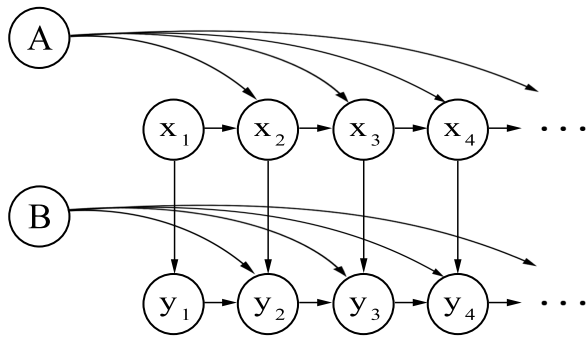}}
	\caption{Graphical representation of GPDM. A and B denote the weights for basis functions.}
	\label{Fig1}
\end{figure}

The Gaussian Process Dynamical Model (GPDM) is primarily employed to analyse the dynamics of latent variables (or low-dimensional embeddings), and it contains a non-linear probabilistic mapping from the latent to observation space,  together with a dynamical model in the latent space \cite{wang2005gaussian}. 
It can be represented as a graphical model as shown  in Fig.~\ref{Fig1}.

We are given training data $\mathbf Y=\lbrack{\mathbf y}_1\dots{\mathbf y}_T\rbrack^\top\in\mathbb{R}^{T\times D}$, which contains $T$ observations $\mathbf{y}_n\in\mathbb{R}^D$, as described in section \ref{probdef}. Let $\mathbf X=\lbrack{\mathbf x}_1\dots{\mathbf x}_T\rbrack^\top\in\mathbb{R}^{T\times Q}$ be a matrix of the latent variables $\mathbf{x}_n\in\mathbb{R}^Q$, with $Q \ll D$. We consider the following model with first-order Markovian dynamics 
\begin{equation}
	\label{eq:linearDM}
	\begin{aligned}
		\mathbf{x}_n &= f({\bf x}_{n-1};{\bf A})+ \mathbf{n}_{x,n}=\sum_{i=1}^{K} \mathbf{a}_i \phi_i(\mathbf{x}_{n-1}) + \mathbf{n}_{x,n}:={\bf A}^\top\boldsymbol{\phi}(\mathbf{x}_{n-1})+ \mathbf{n}_{x,n}
		\\
		\mathbf{y}_n &= g({\bf x}_n;{\bf B})+ \mathbf{n}_{y,n}=\sum_{j=1}^{M} \mathbf{b}_j \psi_j(\mathbf{x}_{n}) + \mathbf{n}_{y,n}:={\bf B}^\top\boldsymbol{\psi}(\mathbf{x}_{n})+ \mathbf{n}_{y,n},
	\end{aligned}
\end{equation}
in which the rows of $\mathbf A=\lbrack{\mathbf a}_1\dots{\mathbf a}_K\rbrack^\top\in\mathbb{R}^{K\times Q}$ and $\mathbf B=\lbrack{\mathbf b}_1\dots{\mathbf b}_M\rbrack^\top\in\mathbb{R}^{M\times D}$ are weights, while \{$\phi_i(\mathbf{x})\}$ and $\{\psi_j(\mathbf{x})\}$ are basis functions and $\mathbf{n}_{y,n}$, $\mathbf{n}_{x,n}$ are zero-mean Gaussian processes representing noise. The models above are essentially generalised linear models.

The use of explicit basis functions with some user specified numbers $K$ and $M$ and associated weights $\mathbf A$ and $\mathbf B$ is a nuisance that can be eliminated using a Bayesian approach that marginalises out both using Gaussian priors. We first explain the original approach, in which ${\bf B}$ is integrated out by placing  independent isotropic priors over its  columns $\widehat{\mathbf{b}}_d\in\mathbb{R}^M$, $d=1,\hdots,D$: 
$p(\widehat{\mathbf{b}}_d\mid w_d)={\cal N}\left({\bf 0},w_d^{-2}{\bf I}\right)$. The model for each column $\widehat{\mathbf{y}}_d$ of ${\bf Y}$ (the $d$-th attribute) is $\widehat{\mathbf{y}}_d= \boldsymbol{\Psi} \widehat{\mathbf{b}}_d+{\bf n}_{y,d}$, in which $\boldsymbol{\Psi}=[\boldsymbol{\psi}(\mathbf{x}_{1})\hdots \boldsymbol{\psi}(\mathbf{x}_{T})]^\top$ is a design matrix and ${\bf n}_{y,d}\sim {\cal N}({\bf 0},w_d^{-2}\sigma_Y^2{\bf I})$ is i.i.d. noise. Here, the variance of the error $w_d^{-2}\sigma_Y^2$ is assumed to take a particular form, namely that it is scaled by $w_d^{-2}$ \cite{grochow2004style}. The distribution over $\widehat{\mathbf{y}}_d$ is therefore $p(\widehat{\mathbf{y}}_d\mid {\bf X},\widehat{\mathbf{b}}_d,w_d,\sigma_Y)={\cal N}\left(\boldsymbol{\Psi} \widehat{\mathbf{b}}_d,w_d^{-2}\sigma_Y^2{\bf I}\right)$. 
Standard conditional results applied to the Gaussians $p(\widehat{\mathbf{y}}_d\mid {\bf X},\widehat{\mathbf{b}}_d,w_d,\sigma_Y)$ and $p(\widehat{\mathbf{b}}_d\mid w_d)$ allow us to eliminate $\widehat{\mathbf{b}}_d$ in order to obtain $p(\widehat{\mathbf{y}}_d\mid {\bf X},w_d,\sigma_Y)$ 
\begin{equation}\label{yhatd}
p(\widehat{\mathbf{y}}_d\mid {\bf X},w_d,\sigma_Y)={\cal N}\left({\bf 0},w_d^{-2}{\bf K}_Y\right), \quad {\bf K}_Y=\boldsymbol{\Psi} \boldsymbol{\Psi}^\top +\sigma_Y^2{\bf I},
\end{equation}
in which ${\bf K}_Y$ is an unscaled kernel matrix. 

The kernel matrix ${\bf K}_Y$ can be generated  by an equivalent kernel function, representing covariances  between the components $\widehat{y}_{d,n}$ of $\widehat{\bf y}_d$, i.e,  $\mbox{cov}({y}_{d,n},{y}_{d,n'})=w_d^{-2}k_{Y}({\bf x}_n,{\bf x}_{n'}|\boldsymbol{\theta})$. For example, we can use the squared exponential
\begin{equation}
	\begin{aligned}
		k_{Y}({\bf x}_n,{\bf x}_{n'}|\boldsymbol{\theta}_Y)=\theta_{Y,1}\exp \left(-\frac{\theta_{Y,2}}{2}\lVert \mathbf{x}_n-\mathbf{x}_{n'}\rVert^2\right)+\theta_{Y,3}^{-1}\delta({\bf x}_n-{\bf x}_{n'}),
	\end{aligned}
\end{equation}
in which  $\boldsymbol{\theta}_Y=\left\{\theta_{Y,1}, \theta_{Y,2},\theta_{Y,3}\right\}$ and  $\theta_{Y,3}^{-1}=\sigma_Y^2$. Note that we absorb the noise into the kernel using the delta function $\delta({\bf x}_n-{\bf x}_{n'})$. The complete joint likelihood is then given by, after using the properties of the trace $\text{tr}(\cdot)$ of a matrix 
\begin{equation}
	\label{latent mapping}
	\begin{aligned}
		p(\mathbf{Y} \mid \mathbf{X},\boldsymbol{\theta}_Y,\mathbf{W})= \prod_{d=1}^{D} p(\widehat{\mathbf{y}}_d\mid {\bf X},\boldsymbol{\theta}_Y,w_d)= \frac{|\mathbf{W}|^{T}}{\sqrt{(2\pi)^{TD} |\mathbf{K}_Y|^D}} \exp\left\{-\frac{1}{2} \text{tr} \Big( \mathbf{K}_Y^{-1}  \mathbf{Y} \mathbf{W}^2 \mathbf{Y}^{\top} \Big) \right\},
	\end{aligned} 
\end{equation}
in which $\mathbf{W}=\operatorname{diag}\left(w_{1}, \hdots, w_{D}\right)$

In our modification, we place a  general matrix Gaussian prior  over $\B$
\begin{equation}\label{priorW}
p(\B\mid \K_D,\K_M) =  
\frac{1}{\sqrt{(2\pi)^{MD} |\K_D|^{M} |\textbf{K}_M|^{D}} } \exp\left\{-\frac{1}{2} \mathrm{tr}[\textbf{K}_M^{-1} \B^T\K_D^{-1} \B ] \right\},
\end{equation}
in which  $\textbf{K}_M\in \mathbb{R}^{M\times M}$  and $\K_D\in \mathbb{R}^{D\times D}$ are the row and column covariance matrices.  Marginalising over $\B$  yields 
\begin{equation}\label{equ:baye f}
\begin{aligned}
p\left(\y_n \mid \x_n,\K_M,\K_D,\sigma_Y\right) = \mathcal{N} \left(\textbf{0}, \boldsymbol{\psi}^T(\x_n)\K_M \boldsymbol{\psi}(\x_n)\K_D  + \sigma_Y^2 \I \right).
\end{aligned} \end{equation}
$\textbf{K}_{M}$ is  positive semidefinite (p.s.d.) and therefore possesses a unique p.s.d. square root, so that $\boldsymbol{\psi}^T(\x_n)\K_M \boldsymbol{\psi}(\x_n) $ defines a kernel 
\begin{equation}\label{link}
	 \langle \widetilde{\boldsymbol{\psi}}(\x_n), \widetilde{\boldsymbol{\psi}}(\x_{n'}) \rangle:= \langle {\boldsymbol{\psi}}(\x_n), {\boldsymbol{\psi}}(\x_{n'}) \rangle_{\K_M},
\end{equation}
in which $\widetilde{\boldsymbol{\psi}}(\x_n)=\sqrt{\textbf{K}_{M}}\boldsymbol{\psi}(\x_n)$, $\langle \cdot,\cdot \rangle$ denotes the standard Euclidean  inner product and  $\langle \cdot,\cdot \rangle_{\K_M }$ is an inner product weighted by $\textbf{K}_{M}$. Notice that our approach models the $\y_n$ directly, as opposed to the original approach in which the columns of ${\bf Y}$ are modelled. Using an equivalent kernel $k_{Y}({\bf x}_n,{\bf x}_{n'}|\boldsymbol{\theta}_Y)$  to replace $\boldsymbol{\psi}^T(\x_{n})\K_M \boldsymbol{\psi}(\x_{n'})$ yields the separable model
\begin{equation}\label{GPovery}
\begin{aligned}
\y_n \mid \x_n,\K_D,\boldsymbol{\theta}_Y,\sigma_Y \sim \mathcal{GP} \left(\textbf{0}, k_{Y}({\bf x}_n,{\bf x}_{n'}|\boldsymbol{\theta}_Y)\otimes \K_D  + \delta({\bf x}_n-{\bf x}_{n'})\otimes \sigma_Y^2 \I\right),
\end{aligned} \end{equation}
in which $\mathcal{GP}(\cdot,\cdot)$ denotes a GP with first and second arguments specifying the mean and covariance functions. This leads to the likelihood
\begin{equation}\label{modelIKL}
\begin{aligned}
p\left(\Y \mid \X,\mathbf {L}_Y,\boldsymbol{\theta}_Y,\sigma_Y\right)& =
\frac{1}{\sqrt{(2\pi)^{TD} |\boldsymbol{\Sigma}_Y|^D}} \exp\left\{-\frac{1}{2}  \mathrm{tr}\left(\mbox{vec}(\Y)\mbox{vec}(\Y)^{\top}\boldsymbol{\Sigma}_Y^{-1} \right) \right\}, \\
 \boldsymbol{\Sigma}_Y&=\K_Y\otimes \mathbf {L}_Y\mathbf {L}_Y^{\top}+\sigma_Y^2{\bf I},
\end{aligned}
\end{equation}
in which $\K_Y$ is generated by $k_{Y}({\bf x}_n,{\bf x}_{n'}|\boldsymbol{\theta}_Y)$, $n,n'=1,\hdots,T$. Since $\K_D$ is p.s.d., the correlations across dimensions of ${\bf y}_n$ can be modelled indirectly  using a full-rank or low-rank Cholesky decomposition $\mathbf {K}_D =\mathbf {L}_Y\mathbf {L}_Y^{\top}$, 
where $\mathbf {L}_Y$ is lower triangular.

For the latent mapping, the original GPDM places independent isotropic priors over the  columns $\widehat{\mathbf{a}}_q\in\mathbb{R}^K$, $q=1,\hdots,Q$, of ${\bf A}$, this time using standard normals, $p(\widehat{\mathbf{a}}_q)={\cal N}\left({\bf 0},{\bf I}\right)$. In our formulation, we use the equivalent of (\ref{priorW}) for $\A$
\begin{equation}\label{priorW}
p(\A\mid \K_Q,\K_K) =  
\frac{1}{\sqrt{(2\pi)^{KQ} |\K_Q|^{K} |\textbf{K}_K|^{Q}} } \exp\left\{-\frac{1}{2} \mathrm{tr}[\textbf{K}_K^{-1} \A^T\K_Q^{-1} \A ] \right\}
\end{equation}
in which $\textbf{K}_K\in \mathbb{R}^{K\times K}$  and $\K_Q\in \mathbb{R}^{Q\times Q}$. The model for ${\bf x}_{n}$ is autoregressive so that the distribution over  ${\bf x}_1$ has to be accounted for separately, given that ${\bf x}_0$ is not known.  GPDM assumes the  model  $\widehat{\mathbf{x}}_{q\backslash 1}= \boldsymbol{\Phi}_{\backslash  T} \widehat{\mathbf{a}}_q+{\bf n}_{x,q}$ 
for each column $\widehat{\mathbf{x}}_{q\backslash 1}$ of ${\bf X}_{\backslash  1}$, defined as the matrix  $\mathbf{X}$ excluding the first row. Here, $\boldsymbol{\Phi}_{\backslash  T} =[\boldsymbol{\phi}(\mathbf{x}_{1})\hdots \boldsymbol{\phi}(\mathbf{x}_{T-1})]^\top$ is a design matrix that excludes the last row and ${\bf n}_{x,q}\sim {\cal N}({\bf 0},\sigma_X^2{\bf I})$ is i.i.d. noise with variance $\sigma_X^2$.  The distribution over $\widehat{\mathbf{x}}_q$ is therefore $p(\widehat{\mathbf{x}}_{q\backslash 1}\mid \widehat{\mathbf{a}}_q)={\cal N}\left(\boldsymbol{\Phi}_{\backslash  T}\widehat{\mathbf{a}}_q,\sigma_X^2{\bf I}\right)$ 
and integrating out ${\bf a}_q$ as before leads to $p(\widehat{\mathbf{x}}_{q\backslash 1}\mid \sigma_X)={\cal N}\left({\bf 0},{\bf K}_{X\backslash  T}\right)$, 
in which ${\bf K}_{X\backslash  T}=\boldsymbol{\Phi}_{\backslash  T}\boldsymbol{\Phi}_{\backslash  T}^\top +\sigma_X^2{\bf I}$ is a kernel matrix that can be generated by any equivalent kernel $k_{X}({\bf x}_n,{\bf x}_{n'}|\boldsymbol{\theta}_X)$, $n,n'=1,\hdots,T-1$, with hyperparameters $\boldsymbol{\theta}_X$. The joint likelihood is then given by
\begin{equation}
	\label{latent mapping b}
	\begin{aligned}
		p(\mathbf{X} \mid \boldsymbol{\theta}_X)&= p({\bf x}_1)\prod_{q=2}^{Q} p(\widehat{\mathbf{x}}_{q\backslash 1}\mid \boldsymbol{\theta}_X)\\&=
		\frac{p({\bf x}_1)}{\sqrt{(2\pi)^{(T-1)Q} |\mathbf{K}_{X\backslash  T}|^Q}} \exp\left\{-\frac{1}{2} \text{tr} \Big( \mathbf{K}_{X\backslash  T}^{-1}  \mathbf{X}_{\backslash  1} \mathbf{X}_{\backslash  1}^{\top} \Big) \right\},
	\end{aligned} 
\end{equation}

In our modification, we instead marginalise over $\A$ and define an equivalent kernel $k_{X}({\bf x}_{n-1},{\bf x}_{{n'-1}}|\boldsymbol{\theta}_X)$ to replace  $\boldsymbol{\phi}^T(\x_{n-1})\K_K \boldsymbol{\phi}(\x_{n'-1})$, yielding 
\begin{equation}
\x_n \mid \x_{n-1},\K_Q,\boldsymbol{\theta}_X,\sigma_X \sim \mathcal{GP} \left(\textbf{0}, k_{X}({\bf x}_{n-1},{\bf x}_{{n'-1}}|\boldsymbol{\theta}_X)\otimes \K_Q  + \delta({\bf x}_{n-1}-{\bf x}_{n'-1})\otimes \sigma_X^2 \I \right)
\end{equation}
and the likelihood function
\begin{equation}\label{modelIKLx}
\begin{aligned}
p\left(\X \mid \mathbf {L}_X,\boldsymbol{\theta}_X,\sigma_X\right) &=
\frac{p({\bf x}_1)}{\sqrt{(2\pi)^{(T-1)Q} |\boldsymbol{\Sigma}_X|^Q}} \exp\left\{-\frac{1}{2}  \mathrm{tr}\left(\mbox{vec}(\mathbf{X}_{\backslash  1})\mbox{vec}(\mathbf{X}_{\backslash  1})^{\top}\boldsymbol{\Sigma}_X^{-1}  \right) \right\},\\
\boldsymbol{\Sigma}_X&=\K_{X\backslash  T}\otimes \mathbf {L}_X\mathbf {L}_X^{\top}+\sigma_X^2{\bf I},
\end{aligned}
\end{equation}
with ${\bf K}_{X\backslash  T}$ generated by $k_{X}({\bf x}_n,{\bf x}_{n'}|\boldsymbol{\theta}_X)$, $n,n'=1,\hdots,T-1$ and $\mathbf {L}_X$ being the equivalent of $\mathbf {L}_Y$. ${\bf x}_1$ can be modelled by an isotropic Gaussian prior but is irrelevant as far as minimising the negative log posterior is concerned.  We note that neither (\ref{latent mapping b}) nor (\ref{modelIKLx}) is Gaussian. 

For dynamical problems, common kernels such as the squared exponential or Mat\'{e}rn are often used in linear combinations with a linear kernel
\begin{equation}
	\begin{aligned}
		k_{X}({\bf x}_n,{\bf x}_{n'}|\boldsymbol{\theta}_X)=\theta_{1,X} \exp \left(-\frac{\theta_{2,X}}{2}\left\|\mathbf{x}-\mathbf{x}^{\prime}\right\|^{2}\right)+\theta_{3,X} \mathbf{x}^{T} \mathbf{x}^{\prime},
	\end{aligned}
\end{equation}
in which $\boldsymbol{\theta}_X=\left\{\theta_{1,X}, \theta_{2,X}, \theta_{3,X}\right\}$. We also consider multi-kernels formed by linear combinations of up to six different kernels \cite{zhu2016dynamic}
\begin{equation}
	\begin{aligned}
		k_{X}({\bf x}_n,{\bf x}_{n'}|\boldsymbol{\theta}_X)=\sum_{l=1}^{L} w_{l} k_l({\bf x}_n,{\bf x}_{n'}|\boldsymbol{\theta}_l),
	\end{aligned} 
\end{equation}
in which the $k_l({\bf x}_n,{\bf x}_{n'}|\boldsymbol{\theta}_l)$ denote different kernels with hyperparameters $\boldsymbol{\theta}_l$, and $\boldsymbol{\theta}_X=\{\boldsymbol{\theta}_l\}$.

\subsection{Training via maximum likelihood}\label{Learning}

To learn ${\bf X}$  we need to infer the posterior $p(\mathbf{X} \mid \mathbf{Y})$ according to Bayes' rule, assuming that the hyperparameters have already been inferred 
\begin{equation}
	\label{posterior}
	\begin{aligned}
		p(\mathbf{X} \mid \mathbf{Y}) =\frac{p(\mathbf{X} ,\mathbf{Y})}{p(\mathbf{Y})}\propto p(\mathbf{Y} \mid \mathbf{X})p(\mathbf{X}).
	\end{aligned}
\end{equation}
This can be achieved by approximately sampling from the posterior using Monte Carlo methods, or by using a point estimate, such as maximising the log of the posterior, treating the denominator as a constant. However, a joint maximisation in which the hyperparameters are  simultaneously learned is simpler. To avoid overfitting, we  place inverse priors on the kernel hyperparameters and variances
%
\begin{equation}
p(\boldsymbol{\theta}_j) \propto \prod_{i} \theta_{i,j}^{-1},\quad 
p(\sigma_j^2)\propto \sigma_j^{-2},\quad  j\in\{X,Y\},
\end{equation}
and thus rewrite Eq. \eqref{posterior} as
%
\begin{equation}
	\label{model}
	\begin{aligned}
		p(\mathbf{X}, \boldsymbol{\theta}_X, \boldsymbol{\theta}_Y,\sigma_X,\sigma_Y\mid \mathbf{Y},{\bf L}_X,{\bf L}_Y ) \propto p(\mathbf{Y} \mid \mathbf{X}, {\bf L}_Y,\boldsymbol{\theta}_Y,\sigma_Y) p(\mathbf{X} \mid {\bf L}_X,\boldsymbol{\theta}_X,\sigma_X) \prod_{j}p(\boldsymbol{\theta}_j)p(\sigma_j^2).
	\end{aligned} 
\end{equation}
We now minimize the negative  joint log posterior $-\ln p(\mathbf{X}, \boldsymbol{\theta}_X, \boldsymbol{\theta}_Y,\sigma_X,\sigma_Y\mid \mathbf{Y},{\bf L}_X,{\bf L}_Y )$, ignoring any constants

\begin{equation}
	\label{nlpb}
	\begin{aligned}
		\mathbf{X}^*, \boldsymbol{\theta}_X^*, \boldsymbol{\theta}_Y^*,{\bf L}_X^*,{\bf L}_Y^*,\sigma_X^*,\sigma_Y^*= \;\;&\mbox{arg}\hspace{-0.4in}\min_{\substack{\vspace{1mm}\\ \hspace{-0.15in}\mathbf{X}, \boldsymbol{\theta}_X, \boldsymbol{\theta}_Y,{\bf L}_X,{\bf L}_Y,\sigma_X,\sigma_Y}}\hspace{-0.35in}-\mathcal{L}(\mathbf{X}, \boldsymbol{\theta}_X, \boldsymbol{\theta}_Y,{\bf L}_X,{\bf L}_Y,\sigma_X,\sigma_Y)  \\
		\mathcal{L}(\mathbf{X}, \boldsymbol{\theta}_X, \boldsymbol{\theta}_Y,{\bf L}_X,{\bf L}_Y,\sigma_X,\sigma_Y) =& -\frac{Q}{2} \ln \left|\boldsymbol{\Sigma}_X\right|-\frac{1}{2} \mathrm{tr}\left(\mbox{vec}(\mathbf{X}_{\backslash  1})\mbox{vec}(\mathbf{X}_{\backslash  1})^{\top}\boldsymbol{\Sigma}_X^{-1}  \right)-\frac{D}{2} \ln \left|\boldsymbol{\Sigma}_Y\right|\\
		&-\frac{1}{2} \mathrm{tr}\left(\mbox{vec}(\Y)\mbox{vec}(\Y)^{\top}\boldsymbol{\Sigma}_Y^{-1} \right) -\sum_{i,j} \ln \theta_{i,j}-2\sum_{j} \ln \sigma_{j}.
	\end{aligned}
\end{equation}
${\bf X}$ can be initialised and a value of $Q$ selected using a principal component analysis, selecting the $Q$ features (principal components) with the highest variance (eigenvalue corresponding to the principal direction).

\subsection{Prediction with observations}

To predict future values of the observables $\y$,  we adopt the simple yet effective method of  mean-prediction \cite{wang2007gaussian}, namely we use only the mean prediction in the GP predictive posterior and neglect the variances. Based on the first-order Markovian dynamics, we predict $\mathbf{x}_{n}$ conditioned on $\mathbf{x}_{n-1}$ using Gaussian conditioning rules, as in the original GPDM

\begin{equation}
\begin{aligned}
p&({\mathbf{x}}_{n}\mid {\mathbf{x}}_{n-1},\boldsymbol{\theta}_X,{\bf L}_X)={\cal N}\left(\boldsymbol{\mu}_{X}\left({\mathbf{x}}_{n-1}\right), \boldsymbol{\Lambda}_X\left({\mathbf{x}}_{n-1}\right) \right),\\
&\boldsymbol{\mu}_{X}\left({\mathbf{x}}\right)= \left(\textbf{L}_X\textbf{L}_X^\top \otimes {\bf k}_{X\backslash  T}(\mathbf{x})\right)^\top  \bSigma_X^{-1}\; \text{vec}(\textbf{X}_{\backslash  1}),\\
&\boldsymbol{\Lambda}_X\left({\mathbf{x}}\right) = 
k_{X}({\bf x},{\bf x}|\boldsymbol{\theta}_X)\,\textbf{L}_X\textbf{L}_X^\top -\left(\textbf{L}_X\textbf{L}_X^\top \otimes {\bf k}_{X\backslash  T}(\mathbf{x})\right)^\top  \bSigma_X^{-1}\left(\textbf{L}_X\textbf{L}_X^\top \otimes {\bf k}_{X\backslash  T}(\mathbf{x})\right),\\
\end{aligned}
\label{eq:post Ha}
\end{equation}
in which ${\bf k}_{X\backslash  T}(\mathbf{x})=[k_{X}({\bf x},{\bf x}_{1}|\boldsymbol{\theta}_X),\hdots,k_{X}({\bf x},{\bf x}_{T-1}|\boldsymbol{\theta}_X)]^\top$. To be clear, the  forecasting consists of iteratively using the mean estimate ${\mathbf{x}}_{n-1} = \boldsymbol{\mu}_{X}({\mathbf{x}}_{n-2})$ to estimate  ${\mathbf{x}}_{n}$ via Eq. (\ref{eq:post Ha}).  Similarly, we infer $\mathbf{y}_n$ as the mean of the following posterior predictive distribution using the mean estimate of ${\mathbf{x}}_{n}$
\begin{equation}
\begin{aligned}
&p({\mathbf{y}}_{n}\mid {\mathbf{x}_n},\boldsymbol{\theta}_Y,{\bf L}_Y)={\cal N}\left(\boldsymbol{\mu}_{Y}\left({\mathbf{x}_n}\right), \boldsymbol{\Lambda}_Y\left({\mathbf{x}_n}\right) \right),\\
&\boldsymbol{\mu}_{Y}\left({\mathbf{x}}\right)= \left(\textbf{L}_Y\textbf{L}_Y^\top \otimes {\bf k}_{Y}(\mathbf{x})\right)^\top  \bSigma_Y^{-1}\; \text{vec}(\textbf{Y}),\\
&\boldsymbol{\Lambda}_Y\left({\mathbf{x}}\right) = 
k_{Y}({\bf x},{\bf x}|\boldsymbol{\theta}_Y)\,\textbf{L}_Y\textbf{L}_Y^\top -\left(\textbf{L}_Y\textbf{L}_Y^\top \otimes {\bf k}_{Y}(\mathbf{x})\right)^\top  \bSigma_Y^{-1}\left(\textbf{L}_Y\textbf{L}_Y^\top \otimes {\bf k}_{Y}(\mathbf{x})\right),
\end{aligned}
\label{eq:post Hb}
\end{equation}
in which ${\bf k}_{Y}(\mathbf{x})=[k_{Y}({\bf x},{\bf x}_{1}|\boldsymbol{\theta}_Y),\hdots,k_{Y}({\bf x},{\bf x}_{T}|\boldsymbol{\theta}_Y)]^\top$.

\subsection{Datasets}

We assess the proposed method using two NASA Ames Prognostics Center of Excellence Battery Dataset from Saha and Goebel \cite{saha2007battery} and the Oxford Battery Degradation Dataset of Birkl \cite{birkl2017oxford}. Both datasets record the charging and discharging performance  of Li-ion batteries and are suitable for evaluating algorithms for   state-of-health forecasting.

We used datasets pertaining to seven batteries from two different groups in the NASA dataset, with both groups  subjected to repeated charge-discharge cycles. The batteries  are labeled B0005, B0006, B0007 (group 1), and B0029, B0030, B0031, B0032 (group 2). All batteries were charged using a constant current of 1.5 A until the voltage reached the limit 4.2 V, following which a constant voltage was applied until the current reached 20 mA. The operating temperature for the group 1 tests was 24 °C, while the temperature for the group 2 tests was 43 °C. Batteries in group 1 were discharged at a constant current of 2 A until the voltage fell to 2.7 V, 2.5 V and 2.2 V for B0005, B0006 and B0007, respectively. Batteries in group 2 were discharged at a constant current of 4 A  to 2.0 V, 2.2 V, 2.5 V and 2.7 V for B0029, B0030, B0031 and B0032, respectively. For group 1 batteries, 168 charge-discharge cycles were recorded, whereas for group 2 batteries, 39 cycles were recorded. For each battery, the SOH at the end of the cycling was different. The data in each case consists of temperature, voltage, current and capacity measurements.  Although the data also contains impedance measurements, they  were not used in this study due to the very high  levels of  noise they contain.

The Oxford dataset contains measurements from Li-ion pouch cells, all tested in a thermal chamber at 40 °C. The cells were exposed to a constant-current and constant-voltage charging profile, followed by a drive cycle discharging profile derived from the urban Artemis profile, with measurements taken every 100 cycles. The batteries are labelled OX1, OX2 and OX3. Again, the data contains measurements of temperature, voltage, current and capacity on each cycle.

\subsection{Data preprocessing}
\begin{figure}[h]
	\centering
	\subfigure
	{\label{fig:c}\includegraphics[width=50mm]{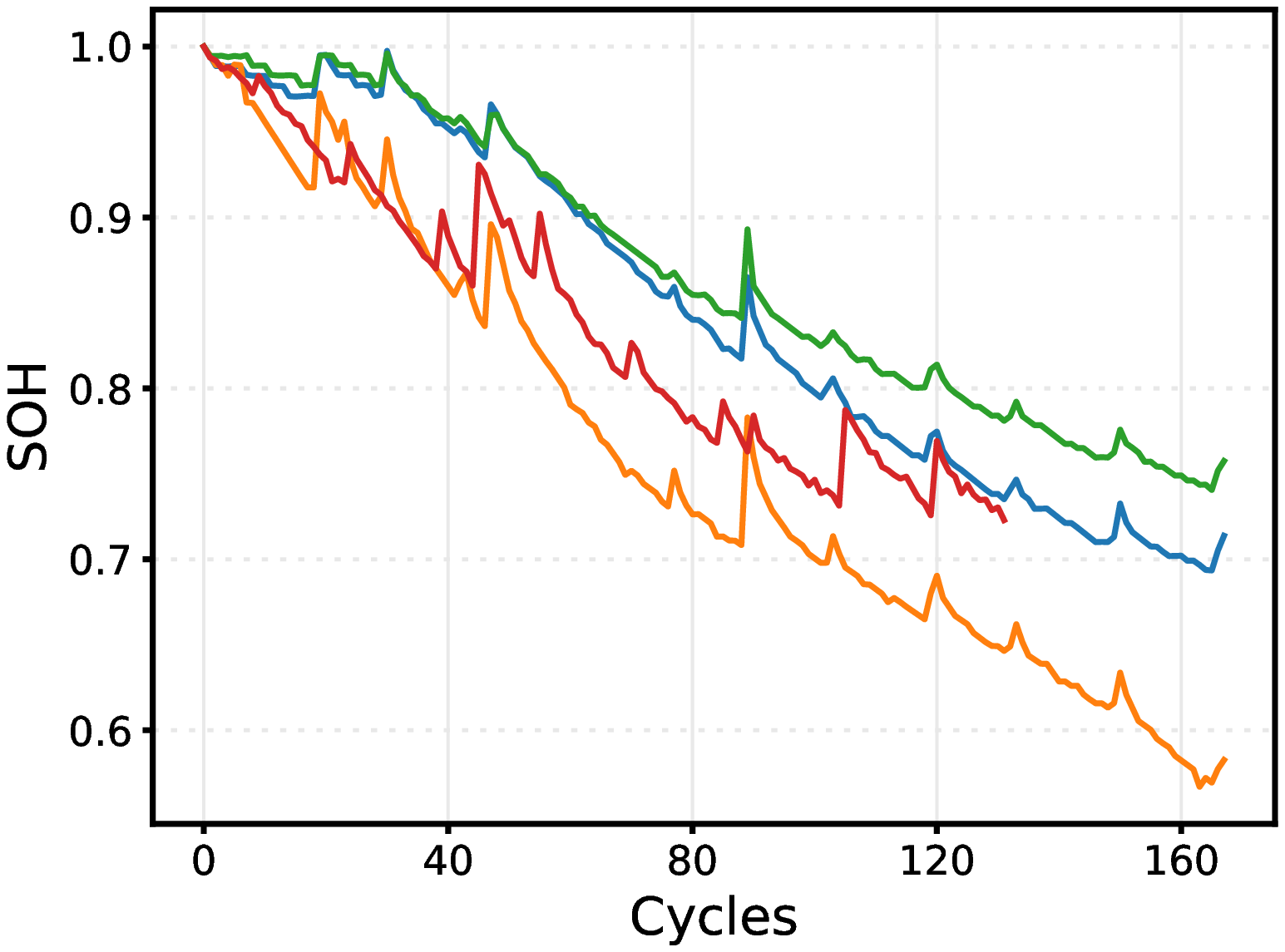}}
	\subfigure
	{\label{fig:c}\includegraphics[width=50mm]{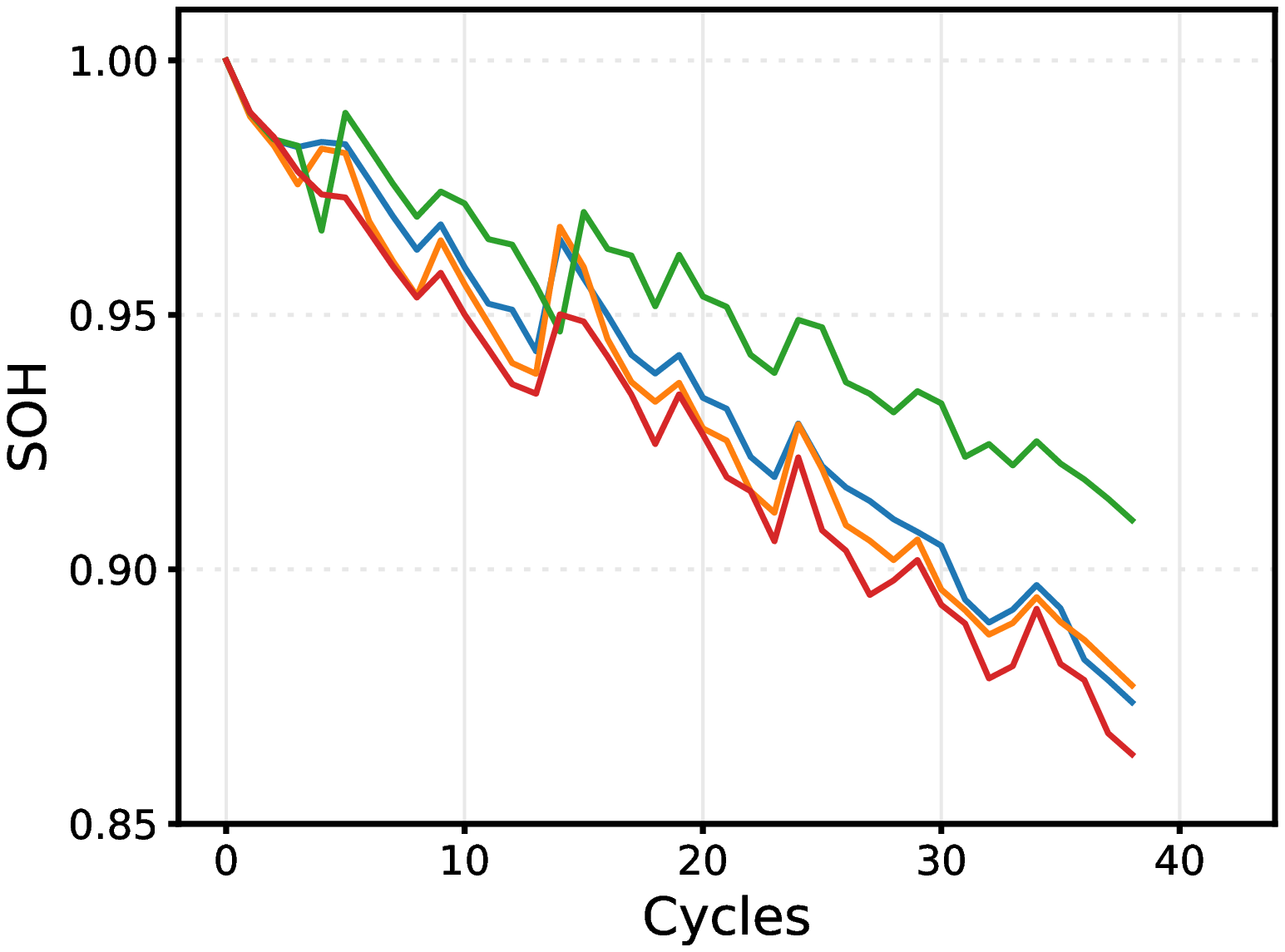}}
	\subfigure
	{\label{fig:c}\includegraphics[width=50mm]{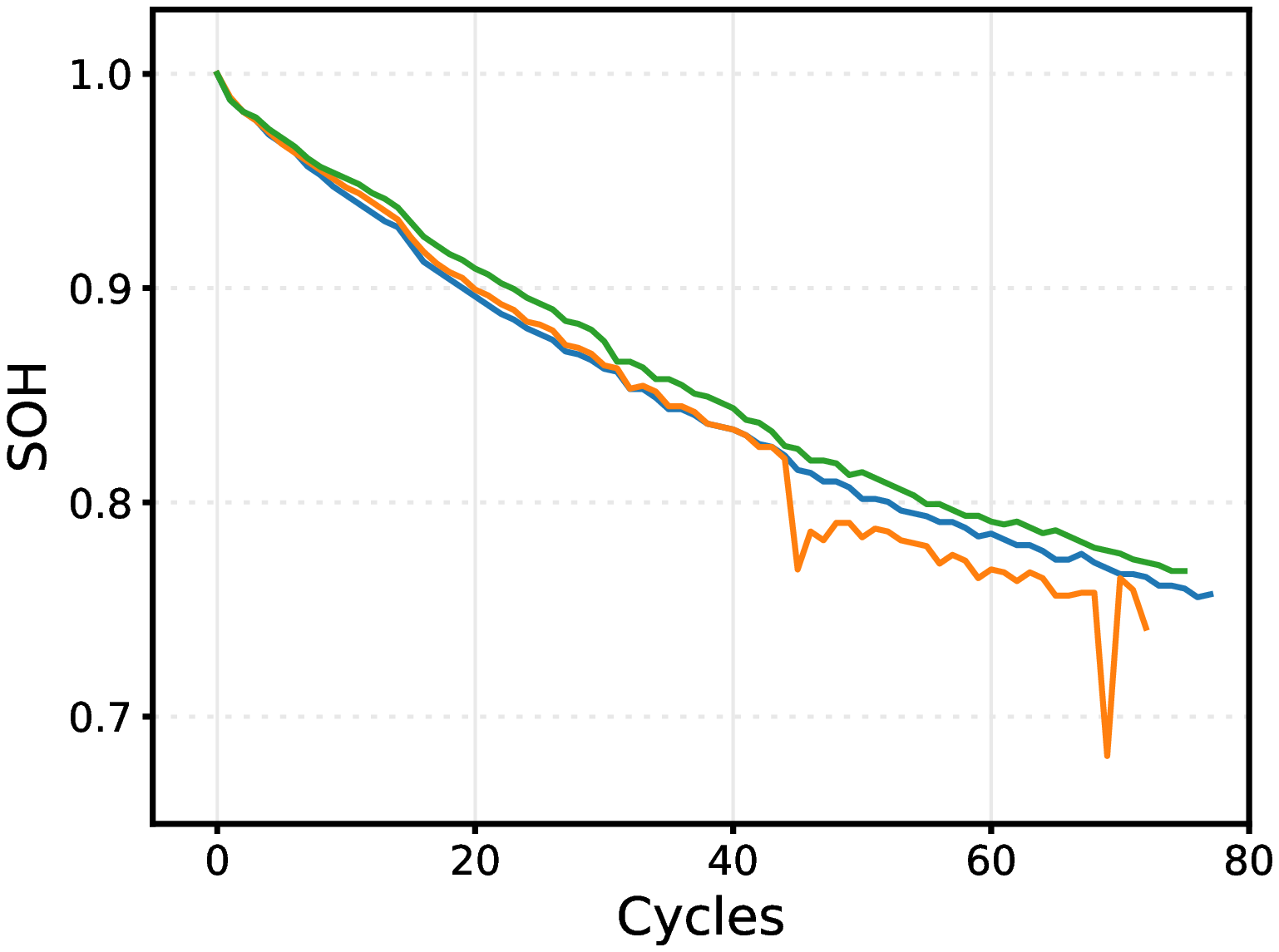}}
	\caption{SOH for 3 groups of batteries. {Left:} B0005, B0006, B0007, and B0018 from the NASA dataset. {Middle:} B0029, B0030, B0031 and  B0032 from the NASA dataset. {Right:} OX1, OX2 and OX3 from the Oxford dataset.}\label{Fig2}
\end{figure}

The raw data was pre-processed as follows:
\begin{enumerate}
	\item {\it Cut off\/}. Voltage and temperature sequences during the discharge cycles were truncated after the voltage reached the threshold (cut-off) value as set in the experiment. 
	\item {\it Interpolation\/}. For the NASA dataset, cubic splines were used  to obtain voltage and temperature data on a fixed grid, with 200 measurements for each cycle. The Oxford dataset did not require interpolation since there are an equal number of measurements  for each cycle.
	\item {\it Rescaling\/}. 
	The capacity was rescaled to define the SOH as described earlier, while the temperature and voltage and other attributes were scaled using a min-max normalisation.

\end{enumerate}
Fig. \ref{Fig2} shows the SOH for the batteries we utilised for the model development and evaluation. 

\section{Results and discussion}

\subsection{Model training and evaluation}

In this section we compare the results of GPDM to those from a number of baseline methods: the GP model of Richardson et al. \cite{richardson2017gaussian}, a CNN, an LSTM, and stacked LSTMs.  A bi-LSTM was found to work best and only results for this architecture are therefore presented. All models were evaluated with 5 different random seeds, and the average errors were calculated. 

We tested the GP models with combinations of 6 different kernel functions,  including the linear, squared-exponential (RBF), polynomial, rational quadratic, and Mat\'{e}rn kernels. Amongst these, a combination of Mat\'{e}rn 3 and  Mat\'{e}rn 5 gave the best performance for the GP model, while a mixed RBF and linear kernel gave the best performance for GPDM. The latent variables ${\bf x}_n$ are embeddings of  the data ${\bf y}_n$. To initialise ${\bf X}$, a PCA on $\Y$ was conducted, and all principal components  were used to define the latent variables ${\bf x}$. Stochastic gradient descent (SGD) was used for the MLE solution. 

The bi-LSTM used two LSTM layers with a tanh activation and a sigmoid recurrent activation, followed by a dense layer of size 64 neurons and a final output layer. Each LSTM layer was followed by a dropout layer of rate 0.2. Various CNN architectures were investigated and  the  best performing  network used two CNN layers followed by two dense layers of sizes 64 and 32 neurons, along with an output layer.  The CNN layers employed padding, strides of 1, and 16 filters each. Pooling was not used since it was found to deteriorate performance. A dropout layer of ratio 0.25 was used after each CNN layer and Relu activation functions were used in the dense layers. Both of the DNNs were training using the ADAM algorithm \cite{kingma2014adam}, with a learning rate of $0.5\times 10^{-4}$ and a decay rate of  $0.5\times 10^{-5}$. Adding further layers to the DNNs did not improve performance, and in fact generally led to worse performance by virtue of the small data sets and therefore inaccurate learning of a higher number of parameters. 

To assess the accuracy we employed the root mean square error (RMSE), defined as
\begin{equation}
\text{RMSE}=\sqrt{\frac{1}{N_t}\sum_{n}(y(n)-y_t(n))^2}
\end{equation}
in which $y(n)$ is the prediction and $y_t(n)$ is the test value at cycle $n$, $N_t$ is the number of test points and the sum is over all test points. We first discuss the results without the use of attributes other than the SOH. 

\subsection{Comparison to baseline methods without additional attributes}

\begin{figure}[h!]
	\centering
	\subfigure
	{\label{fig:c}\includegraphics[width=0.85\linewidth]{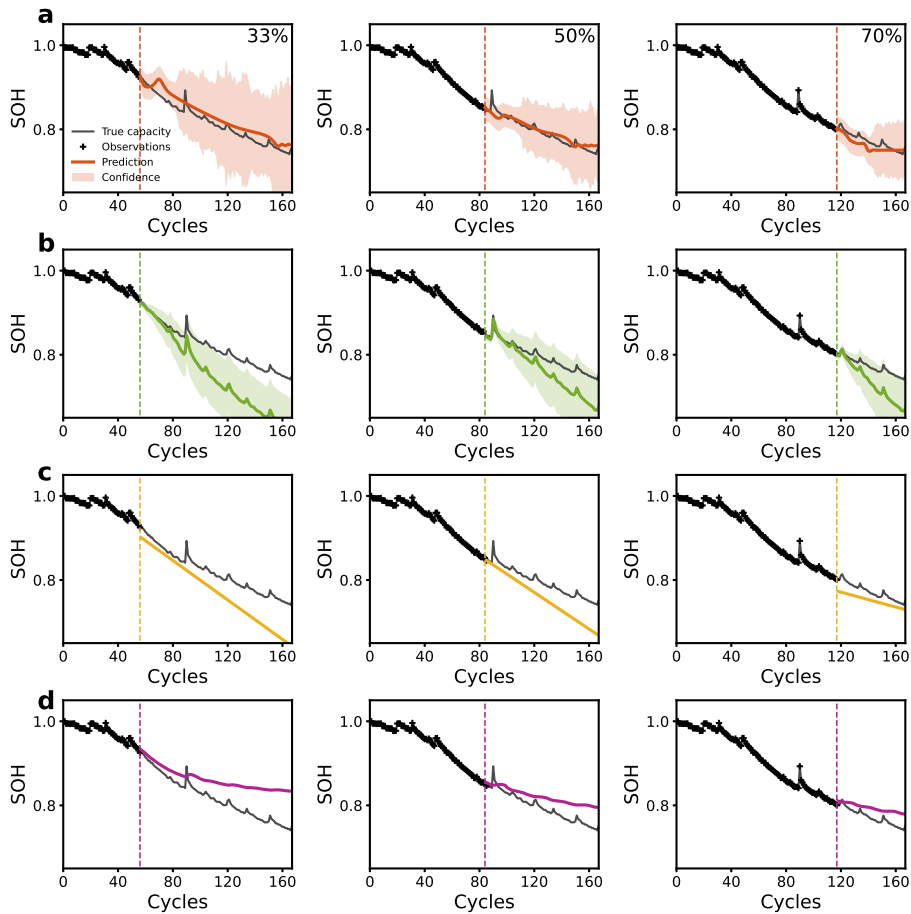}}
	\caption{{State-of-health prediction results for battery B0007.} (a) GPDM (with transfer learning), {(b)} GP, {(c)} CNN, {(d)} bi-LSTM. The figures from left to right correspond to 33\%, 50\% and 70\% training ratios.}
	\label{Fig4}
\end{figure}
In the first experiments, we use the transfer learning described in section \ref{probdef}, with the cycle number,  battery label and SOH as attributes in GPDM. In the  supervised models, the   inputs and outputs  are $\mathbf{z}^{m}_n=[n,m]^\top$ and $\mathbf{y}^{m}_n=\mbox{SOH}^m(n)$, respectively, in which $m\in\{1,\hdots,M\}\subset \mathbb{Z}^+$ is the battery  label. For the unsupervised GPDM, on the other hand, there are only attributes, so that the data takes the form (\ref{encodingd}), i.e.,  $\mathbf{y}_n=[n,m,\mbox{SOH}^m(n)]^\top$.

,

The results from all methods are shown in Fig. \ref{Fig4}, corresponding to the prediction of B0007 with 30\%, 50\% and 70\% of the total data used for training; that is, 30\%, 50\% and 70\% of the B0007 data, while all of the data for B0005 and B0006 was used. The GP models include uncertainty estimates, namely 95\% confidence intervals shown by the the shaded regions in Fig. \ref{Fig4} and defined by
\begin{equation}
	\mu(n)-1.96 \sigma(n) \leq \text{SOH}(n) \leq \mu(n)+1.96 \sigma(n)\tag{12}
\end{equation}
in which $\mu(n)$ and $\sigma(n)$ are the mean and predictive variance at cycle $n$.

The GP model (Fig.~\ref{Fig4}(b)) with transfer learning yields a poor estimate of the rate of decline in SOH, even when presented with 70\% of the data.   Both of the networks (Figs.~\ref{Fig4}(c)\&(d)) essentially fail, which is likely to be  due to the  small volume of training data available.  GPDM (Fig.~\ref{Fig4}(a)) is clearly superior to all of the other methods, achieving high accuracy  for all training point numbers, even 30\%. The overall trend of the SOH is captured well, and the EOL prediction is close to the true value.  Table \ref{Tab1} shows the RMSE values for predictions related to battery B0007 using all methods, together with the RMSE values pertaining to B0005, B0006, B0007, B0029 and B0032. In the latter case, all of the data from batteries in group 2 (B0029-B0032) was used except for that battery under consideration, for which 30, 50 or 70\% was used.

\begin{table}[h]
	\centering
	\scriptsize
	\begin{tabular}{c | c c c c c}
		\hline
		\multicolumn{6}{c}{\bf B0005}\\
		\hline
		\% Train  & \textbf{{GPDM}}  & \textbf{{GPDM}\textsuperscript{\dag}} & GP & CNN & bi-LSTM\\
		\hline
		33     &  \textbf{0.0147}  &  0.0588  & 0.0630  & 0.0313 & 0.0376\\
		50     &  0.0388  &  \textbf{0.0227}  & 0.0324  & 0.0289 & 0.0312\\
		70     &  0.0321  &  0.0640  & 0.0278  & 0.0350 & \textbf{0.0101}\\
		\hline 
		\multicolumn{6}{c}{\bf B0006}\\
		\hline
		33     &  \textbf{0.0189}    & 0.0321  & 0.0499  & 0.0330 & 0.0702 \\
		50     &  0.0458    & 0.0408  & 0.0322  & 0.0319 & \textbf{0.0211} \\
		70     &  \textbf{0.0286}    & 0.0430  & 0.0353  & 0.0311 & \textbf{0.0287} \\
		\hline 
		\multicolumn{6}{c}{\bf B0007}\\
		\hline
		33     &  \textbf{0.0184}  & 0.0800  &  0.0774 & 0.0550  & 0.0479\\
		50     &  \textbf{0.0113}  & 0.0558 &  0.0399 & 0.0441  & 0.0292\\
		70     &  \textbf{0.0128}  & 0.0314  &  0.0473 & 0.0228 & 0.0298\\
		\hline
		\multicolumn{6}{c}{\bf B0029}\\
		\hline
		33      &  \textbf{0.0172}  & 0.0533 & 0.0215 & 0.0207 & 0.0184\\
		50      &  0.0226  & 0.0415 & 0.0280 & \textbf{0.0175} & 0.0195\\
		70      &  \textbf{0.0145}  & 0.0299 & 0.0194 & 0.0190 & 0.0154\\
		\hline
		\multicolumn{6}{c}{\bf B0032}\\
		\hline
		33  &  \textbf{0.0124} & 0.0225 & 0.0201 & 0.0186 & 0.0144\\
		50    &  {\bf 0.0203} & 0.0216 &  \textbf{0.0205} & 0.0229  & \textbf{0.0201}\\
		70   &  \textbf{0.0112}& 0.0145 & 0.0132 & 0.0154 & 0.0132 \\
		\hline
	\end{tabular}
	\caption{RMSE values relating to the SOH predictions using all methods on the NASA data sets. \% Train refers to the percentage of the total data used for training. $\dag$ denotes without transfer learning. }
	\label{Tab1}
\end{table}

\begin{figure}[h!]
	\centering
	\subfigure
	{\label{fig:c}\includegraphics[width=0.85\linewidth]{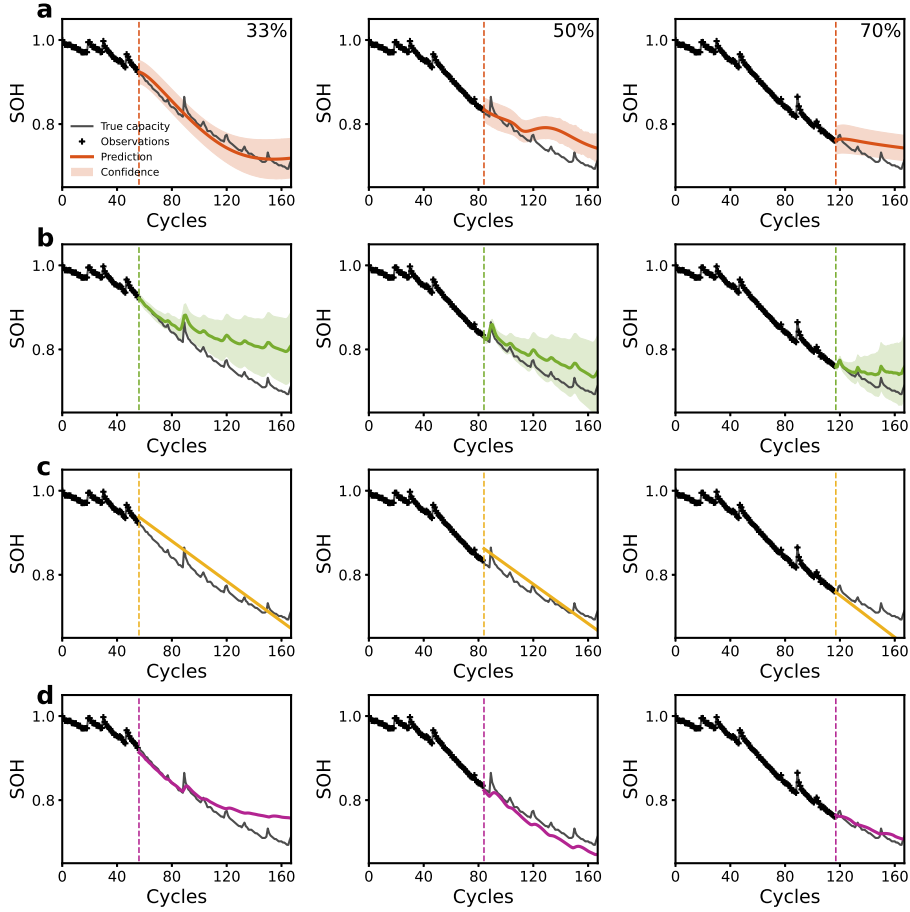}}
	\caption{{State-of-health prediction results for battery B0005.} (a) GPDM (with transfer learning), {(b)} GP, {(c)} CNN, {(d)} bi-LSTM. The figures from left to right correspond to 33\%, 50\% and 70\% training ratios}
	\label{Fig35}
\end{figure}

Also shown in Table \ref{Tab1} is the RMSE without transfer learning for GPDM, from which it is clear that the transfer learning significantly improves the accuracy.  In the case of B0007, the improvement is 77\%, 79\% and  59\% for 30, 50 and 70\% of the training data, respectively. Similar dramatic improvements are seen for B0006, B0029 and B0032. To make the comparisons fair, we used transfer learning for all methods to generate the results in Table \ref{Tab1}, and similar improvements were seen in each case.   As seen in Table \ref{Tab1}, for the group 1 batteries, GPDM (with transfer learning) yields the lowest RMSE in 6 out of 9 cases. Consistent with these result the predictions for group 2 batteries (B0029 and B0032) are superior with GPDM in almost all (5 out of 6) cases. In the case of B0005, GPDM does not perform as well as in the other cases, as seen in Table \ref{Tab1} and Fig. \ref{Fig35}. Nevertheless, the earliest prediction (33\%) is still most accurate with GPDM. Surprisingly, the best performance at 50\% is provided by GPDM without transfer learning, but this result is anomalous.  

\begin{figure}[h!]
	\centering
	\subfigure
	{\label{fig:c}\includegraphics[width=0.85\linewidth]{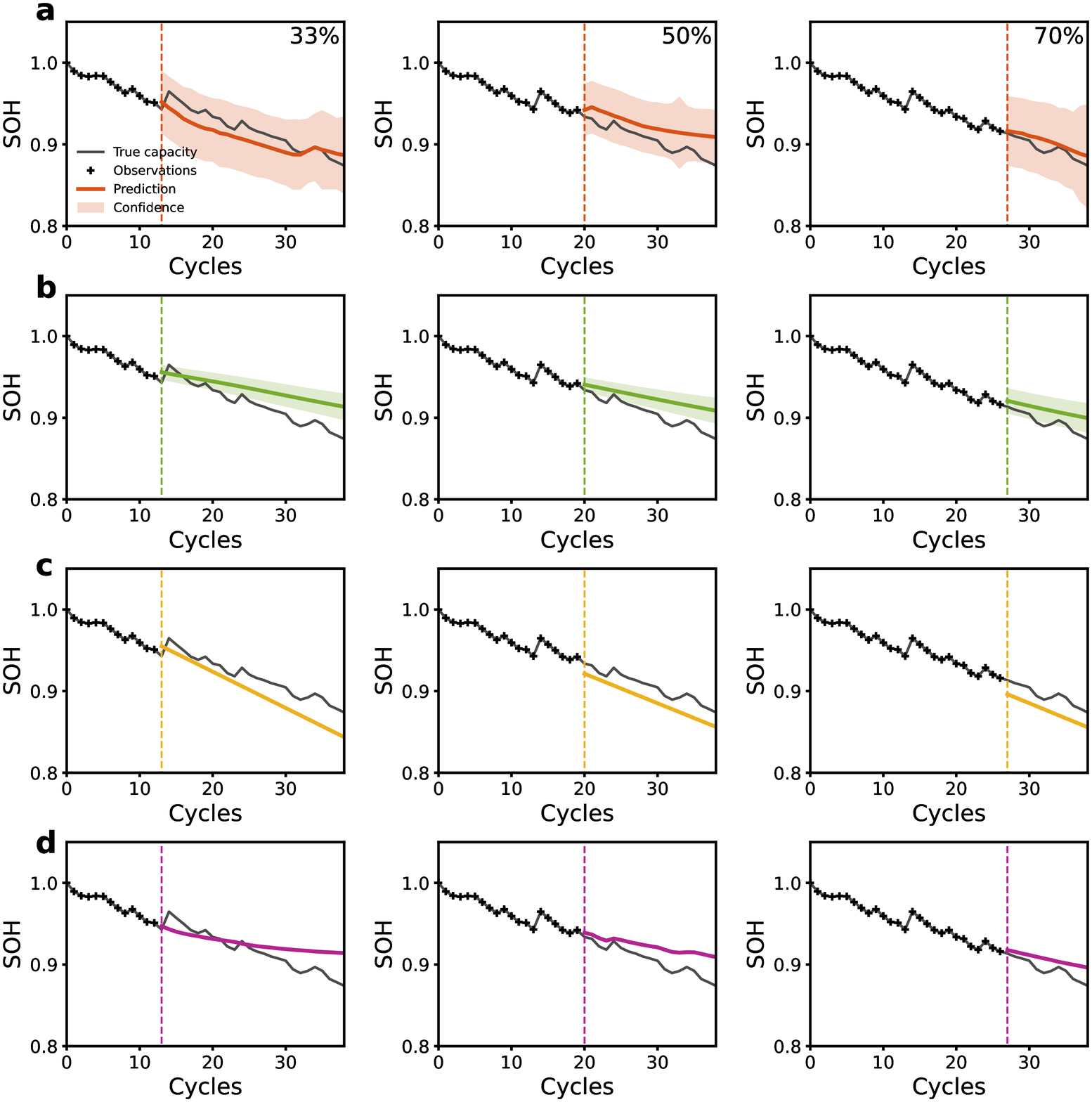}}
	\caption{{State-of-health prediction results for battery B0029.} (a) GPDM (with transfer learning), {(b)} GP, {(c)} CNN, {(d)} bi-LSTM. The figures from left to right correspond to 33\%, 50\% and 70\% training ratios. }
	\label{Fig5}
\end{figure}

The bi-LSTM and CNN also  perform well on B0006 and B0029 when 50\% and 70\% of the data is used, with Bi-LSTM being 23\% more accurate in this case. For B0032, the CNN, Bi-LSTM and GPDM are almost indistinguishable. That the LSTM generally performs better than the CNN is not surprising, since it is designed for sequence problems. The variability of DNNs (model variance) caused by the large numbers of parameters, on the other hand, makes them prone to less reliable overall results unless large volumes of data are available. Fig. \ref{Fig5} shows the predictions for the B0029 battery using all methods.  Here, the superior performance of 
GPDM at the earliest stage of 33\% is again obvious, while at 50\% the   CNN yields the lowest RMSE.

\begin{table}[h!]
	\centering
	\scriptsize
	\begin{tabular}{c | c c c c c}
		\hline
		\multicolumn{6}{c}{\bf OX1}\\
		\hline
		\% Train  & \textbf{{GPDM}} & \textbf{{GPDM}\textsuperscript{\dag}} &  GP & CNN &  bi-LSTM  \\\hline
		33        &  \textbf{0.0093} & 0.0226 &  0.0668 & 0.0199 & 0.0135\\
		50        &  \textbf{0.0135}  & 0.0244 &  0.0207 & 0.0226 & 0.0191\\
		70        &  0.0147 & 0.0263 &  0.0261 & 0.0448  & \textbf{0.0105}\\
		\hline
		\multicolumn{6}{c}{\bf OX3}\\
		\hline
		33        &  \textbf{0.0086}   &  0.0364  &  0.0134 & 0.0207 & 0.0117\\
		50        &  \textbf{0.0099}     & 0.0118  &  0.0123 &  0.0236 & 0.0158\\
		70        &  0.0101   & 0.0089  &  \textbf{0.0061} & 0.0096 & 0.0078\\
		\hline
	\end{tabular}
	\caption{RMSE values relating to the SOH predictions using all methods on the Oxford data set. \% Train refers to the percentage of the total data used for training. $\dag$ denotes without transfer learning. }
	\label{Tab2}
\end{table}

\begin{figure}[h!]
	\centering
	\subfigure
	{\label{fig:c}\includegraphics[width=0.85\linewidth]{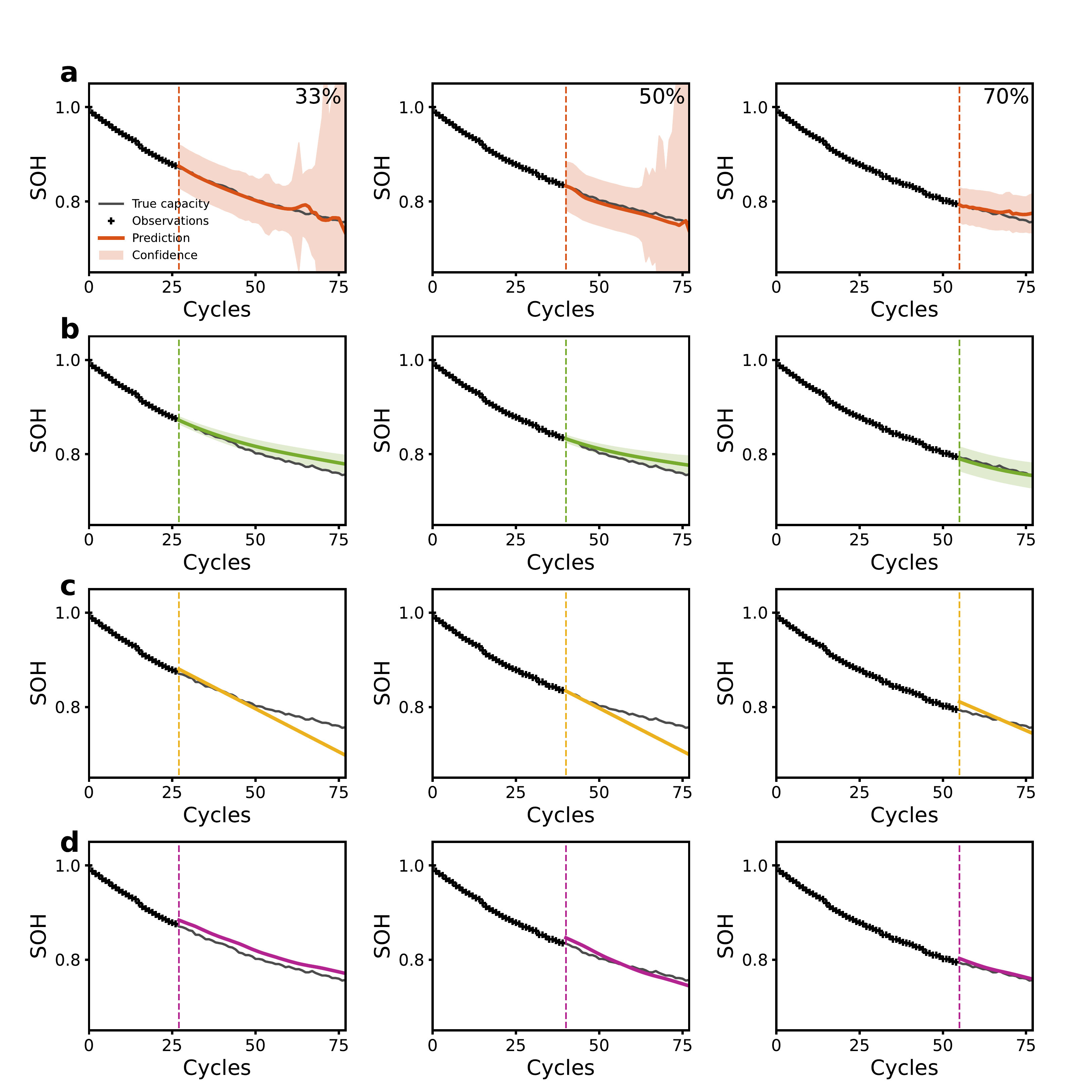}}
	\caption{{State-of-health prediction results for Ox3.} (a) GPDM (with transfer learning), {(b)} GP, {(c)} CNN, {(d)} bi-LSTM. The figures from left to right correspond to 33\%, 50\% and 70\% training ratios. }
	\label{Fig6}
\end{figure}

Finally, Table \ref{Tab2} shows the equivalent RMSE values on the Oxford battery data set, with the transfer learning following the procedure described above. Here again we see that GPDM is generally superior (4 out of 6 cases), with the LSTM and GP occasionally providing the best result (1 out of 6 cases each). As with the NASA data set, the bi-LSTM does well on the short-term predictions corresponding to 70\% of the data for training, with the GP also performing well on Ox3. The early predictions (33\%) are still, however, superior with GPDM, which is respectively 31\% and 44\% more accurate than the next most accurate method (bi-LSTM) on Ox1 and Ox3. For 50\% training data, GPDM is 29\% and 19\% more accurate than the next best method. Fig. \ref{Fig8} shows the predictions corresponding to Ox3 for all methods, clearly showing the superiority of GPDM at 33\%. The excellent performance of bi-LSTM at 70\%  is also evident. 


\subsection{Predictions with additional attributes}

The previous experiments evaluated the predictive power of GPDM for full time-series inference with transfer learning and without attributes other than the SOH.  GPDM was seen to be superior (lower RMSE) when the number of training points was low, corresponding to very early predictions. In other cases (50\% and 70\% of the data used for training) it yielded the lowest RMSE in  50\% of the cases. Overall, GPDM was more accurate in 15 out of 21 (or  in 71\%) of the cases. Although these results are impressive, the main advantage of GPDM is that it allows for the incorporation of an arbitrary number of attributes, which, in theory, can  improve the results. In this section, we therefore incorporate the following attributes
\begin{enumerate}
\item Attribute 1. Temperature at the midpoint of each discharge cycle, $T_m(n)$ 
\item Attribute 2. Voltage at the midpoint of each discharge cycle, $V_m(n)$
\item Attribute 3. Energy delivered by the  battery (unnormalised), given by $I(n)=\int_t V(n)dt$, in which $V(n)$ is the voltage curve (a function of time) for cycle $n$. The integral was evaluated using a  trapezoidal rule
\end{enumerate}
Attributes  1 and 2 will clearly be correlated with SOH, and are two of the most frequently used attributes (although they are usually called features). The (unnormalised) energy delivered by the battery will exhibit a decline as the SOH declines, so again the two will be highly correlated. The choice of these attributes is based on their similarity to those used previously in the literature as well as the information contained in the three data sets under consideration. Explicitly, the data now takes the form (\ref{encodingd}) with explicit attributes
\begin{equation}
\y_n=[n,m,\text{SOH}(n),T_m(n),V_m(n),I(n)]^\top
\end{equation}

The other methods of the previous section cannot be used in this case, for the obvious reason that the attributes will not be known for future cycles. Using known attribute values from the given data set  to perform a multi-cycle lookahead would not lead to a practical algorithm. GPDM, on the other hand, propagates the dynamic hidden variable $\x_n$, from which the observable attributes $\y_n$ (including SOH) can be predicted for an arbitrary number of future cycles. In this case we can compare GPDM with a GPLVM \cite{lawrence2007learning}, which  is able to reconstruct a complete data from partially known features or attributes, a problem that is encountered in many areas, e.g., human motion tracking. This property of GPLVM can be exploited to reconstruct  the complete SOH curve from  33\%, 50\% and 70\% of the data. The GPDM latent space embedding is  initialised in the same way, again using all components from the PCA. We use the linear-plus-RBF kernel and the conjugate gradient method to solve the maximum likelihood problem for both GPDM and GPLVM. 
\begin{figure}[h!]
	\centering
	\subfigure
	{\label{fig:c}\includegraphics[width=0.85\linewidth]{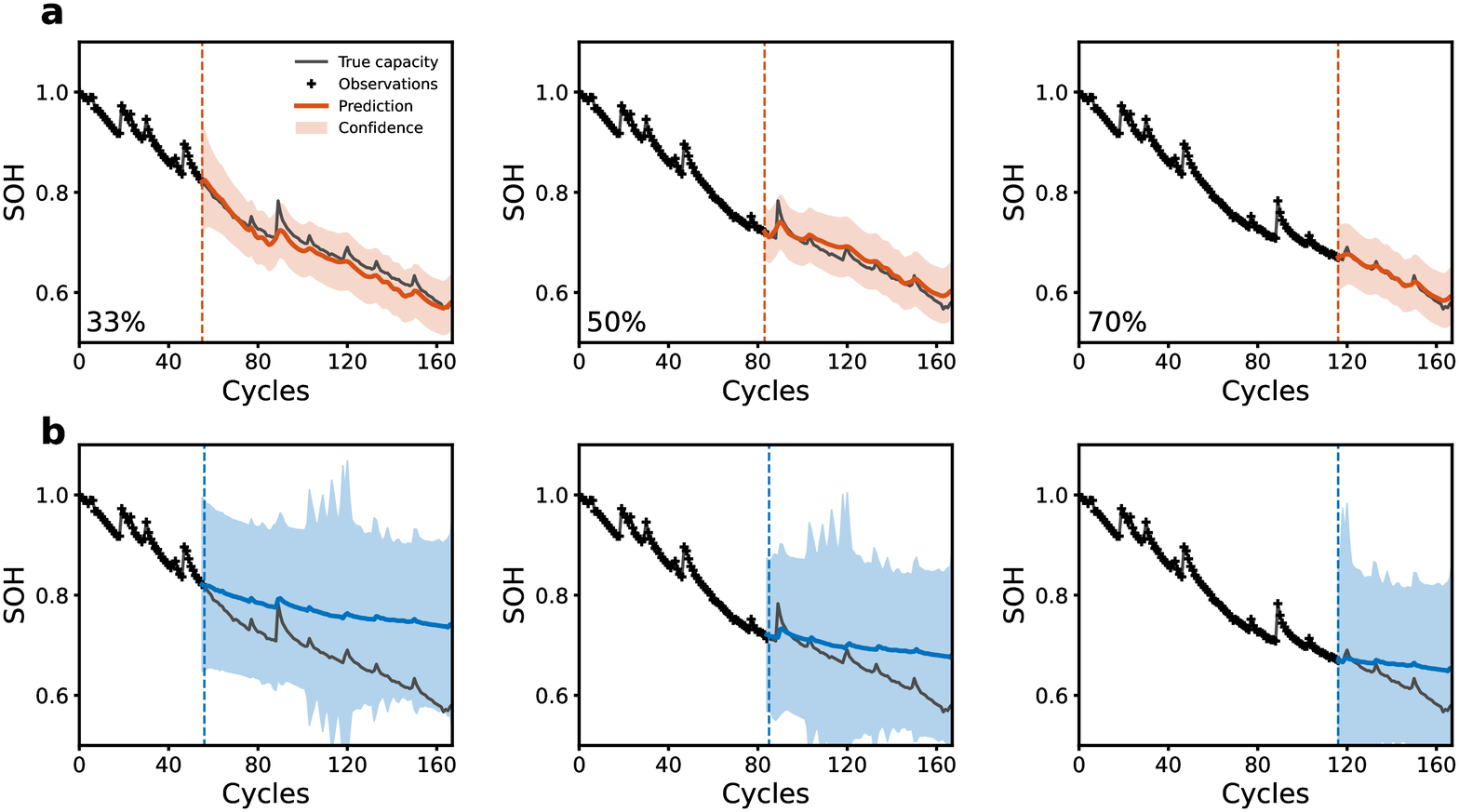}}
	\caption{{State-of-health prediction results for B0006.} (a) GPDM (with transfer learning), (b) GPLVM. The figures from left to right correspond to 33\%, 50\% and 70\% training ratios.}
	\label{Fig7}
\end{figure}

\begin{figure}[h!]
	\centering
	\subfigure
	{\label{fig:c}\includegraphics[width=0.85\linewidth]{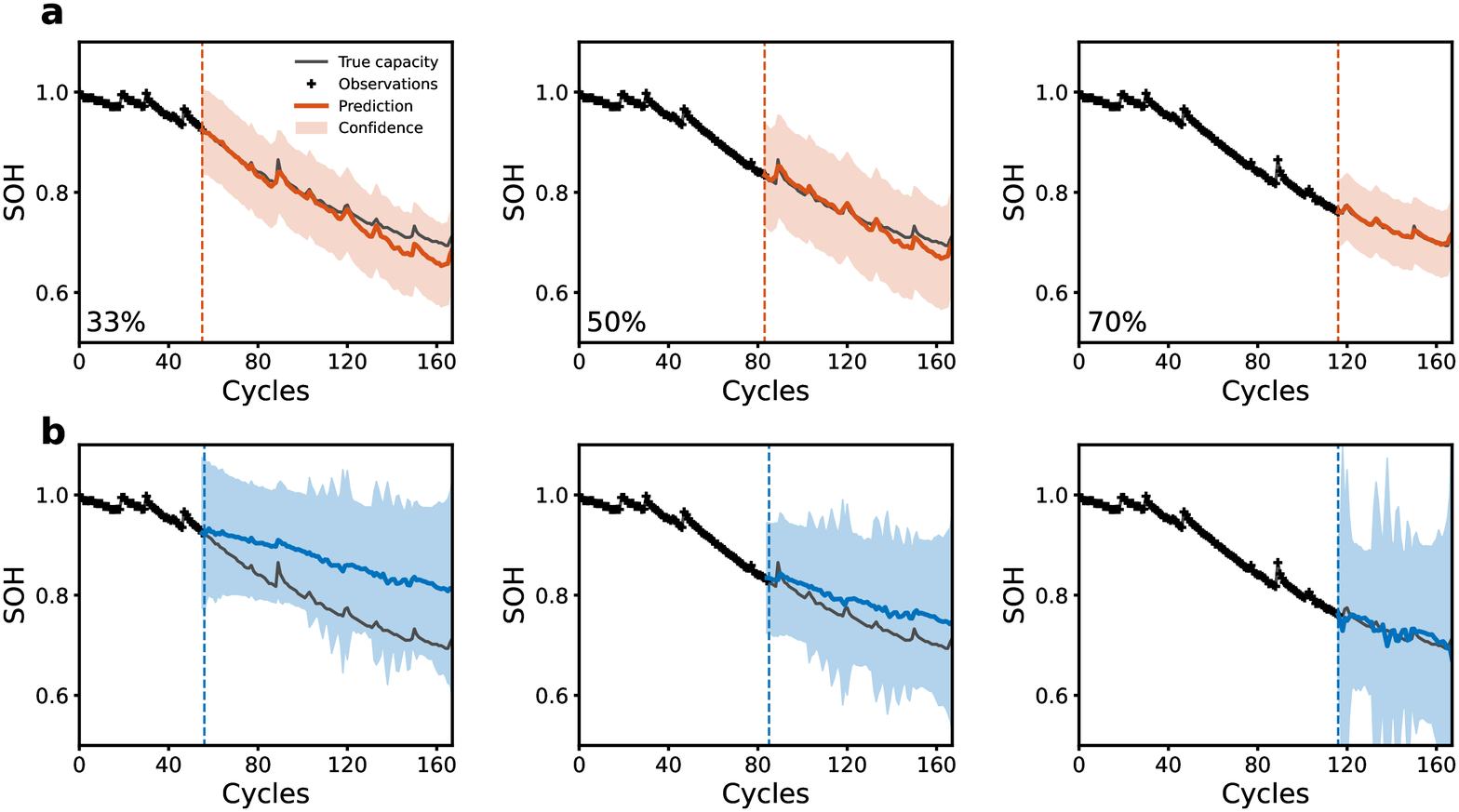}}
	\caption{{State-of-health prediction results for B0005.} (a) GPDM (with transfer learning), (b) GPLVM. The figures from left to right correspond to 33\%, 50\% and 70\% training ratios.}
	\label{Fig8}
\end{figure}
Fig. \ref{Fig7} shows the predictions on B0006 for different training point ratios using GPDM (with and without transfer learning) and GPLVM. It is clear  that GPDM tracks the fluctuations as well as the overall trend of the curve with good accuracy. GPLVM, however, fails both qualitatively and quantitatively. Fig. \ref{Fig8} shows the equivalent results for B0005, which should be compared with Fig. \ref{Fig35} (without additional attributes). We recall that in the latter case, GPLVM did not perform as well as the other methods of the previous section for 50\% and 70\%  training ratios. With attributes included, there is a marked improvement in performance, for all training point ratios. In particular, we point out from Figs.  \ref{Fig7} and  \ref{Fig8} the highly accurate results for 50\% and 70\%.


\begin{table}[h!]
	\centering
	\scriptsize
	\begin{tabular}{c | c c c}
		\hline
		\multicolumn{4}{c}{\bf B0005}\\
		\hline
		\% Train  & \textbf{{GPDM}} & \textbf{{GPDM}\textsuperscript{\dag}} & GPLVM  \cite{lawrence2007learning}
		 \\
		\hline
		33      &  \textbf{0.0152}   & 0.1682   & 0.0884   \\
		50      &  \textbf{0.0134}   & 0.0321   & 0.0348   \\
		70      &  \textbf{0.0029}   & 0.0224   & 0.0141   \\
		\hline
		\multicolumn{4}{c}{\bf B0006}\\
		\hline
		33      &  \textbf{0.0165}    & 0.0410   & 0.0944 \\
		50      &  \textbf{0.0140}    & 0.0325  & 0.0535 \\
		70      &  \textbf{0.0065}    & 0.0166 & 0.0432 \\
		\hline
		\multicolumn{4}{c}{\bf B0007}\\
		\hline
		33      &  \textbf{0.0289}    & 0.0824    &{0.0343} \\
		50      &  \textbf{0.0186}    & 0.0428    & {0.0329} \\
		70      &  \textbf{0.0187}    & 0.0284    & 0.0292 \\
		\hline
		\multicolumn{4}{c}{\bf B0029}\\
		\hline
		33      &  \textbf{0.0176}   & 0.0424  & 0.0378 \\
		50      &  \textbf{0.0240}   & 0.0436  & 0.0389 \\
		70      &  \textbf{0.0050}   & 0.0214  & 0.0203 \\
		\hline
		\multicolumn{4}{c}{\bf B0032}\\
		\hline
		33      &  \textbf{0.0278}   & 0.0284  & 0.0391 \\
		50      &  \textbf{0.0136}   & 0.0342  & 0.0414 \\
		70      &  \textbf{0.0107}   & 0.0223  & 0.0193 \\
		\hline
		\multicolumn{4}{c}{\bf OX1}\\
		\hline
		33      &  \textbf{0.0050}   & 0.0158  & 0.0603 \\
		50      &  \textbf{0.0050}   & 0.0061  & 0.0409 \\
		70      &  \textbf{0.0021}   & 0.0033  & 0.0218 \\
		\hline
		\multicolumn{4}{c}{\bf OX3}\\
		\hline
		33      &  \textbf{0.0099}  & 0.0439  & 0.0630 \\
		50      &  0.0086  & \textbf{0.0056}  & 0.0409 \\
		70      &  \textbf{0.0039}   & \textbf{0.0033} & 0.0213 \\
		\hline
	\end{tabular}
	\caption{RMSE values relating to the SOH predictions using GPDM and GPLVM on all data sets. \% Train refers to the percentage of the total data used for training. $\dag$ denotes without transfer learning. }
	\label{Tab3}
\end{table}

\begin{figure}[h!]
	\centering
	\subfigure
	{\label{fig:c}\includegraphics[width=0.85\linewidth]{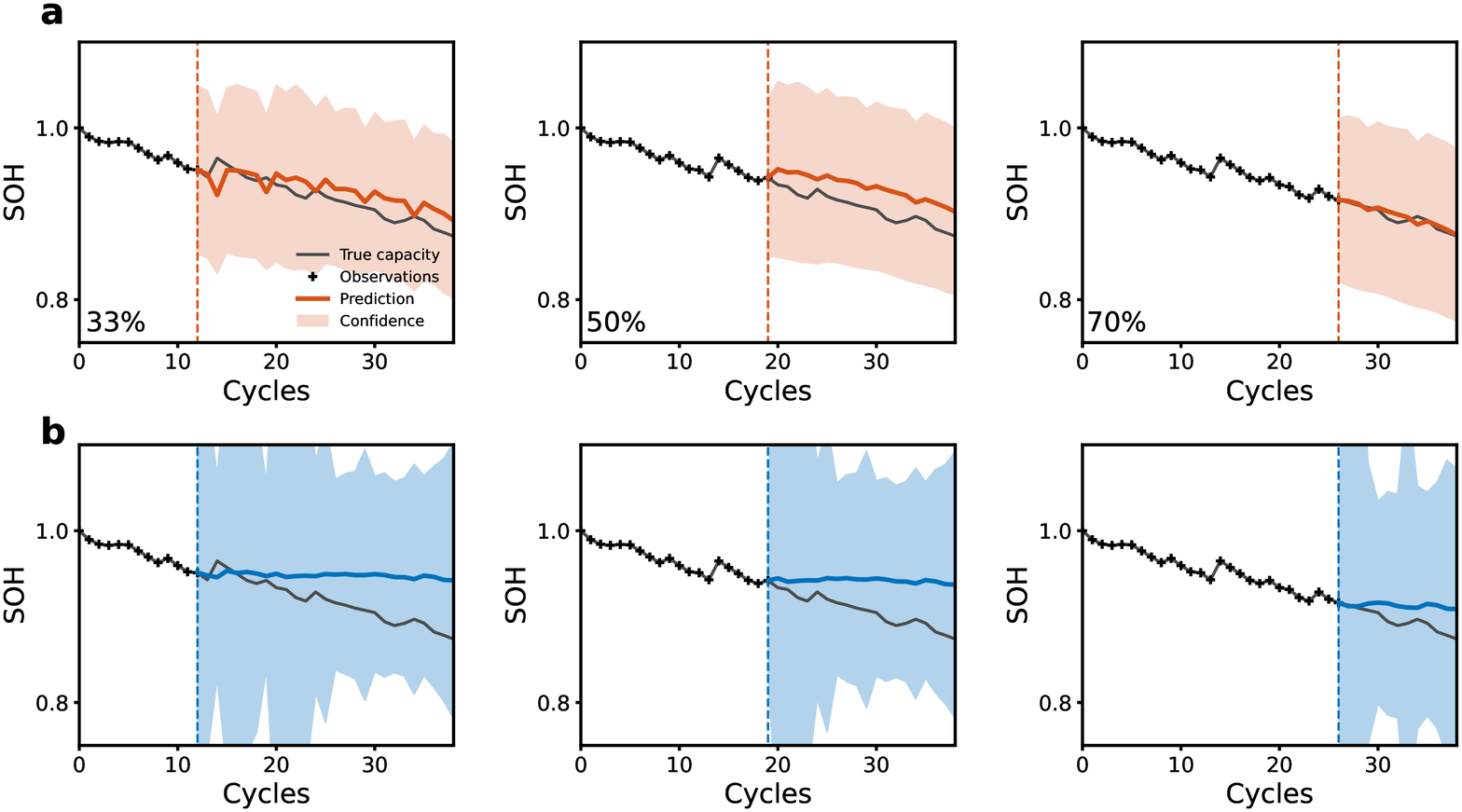}}
	\caption{{State-of-health prediction results for B0029.} (a) GPDM (with transfer learning), (b) GPLVM. The figures from left to right correspond to 33\%, 50\% and 70\% training ratios.}
	\label{Fig9}
\end{figure}

Table \ref{Tab3} shows the RMSE values on all three data sets, which are to be  compared with the values in Tables \ref{Tab1}  and \ref{Tab2}. This comparison shows that in 14 out of 21 cases (76\%), the RMSE values are lower when additional attributes are included. In the case of B0029 (Fig. \ref{Fig9}) at a 33\% training ratio the RMSE values are almost identical, while at 50\% the value is slightly higher (by 6\%). In particular, and as already noted above for B0005 and B0006, for 50\% and 70\%, there is a vast improvement in the RMSE in the vast majority of cases (11 out of 14). For 70\%, the improvement is 71\% (B0005), 77\% (B0006), 66\% (B0029), 4\% (B0032), 80\% (Ox1) and 36\% (Ox3). For 50\%, the improvement is 41\% (B0005), 34\% (B0006), 32\% (B0032), 63\% (Ox1) and 13\% (Ox3). 

\begin{figure}[h!]
	\centering
	\subfigure
	{\label{fig:c}\includegraphics[width=0.85\linewidth]{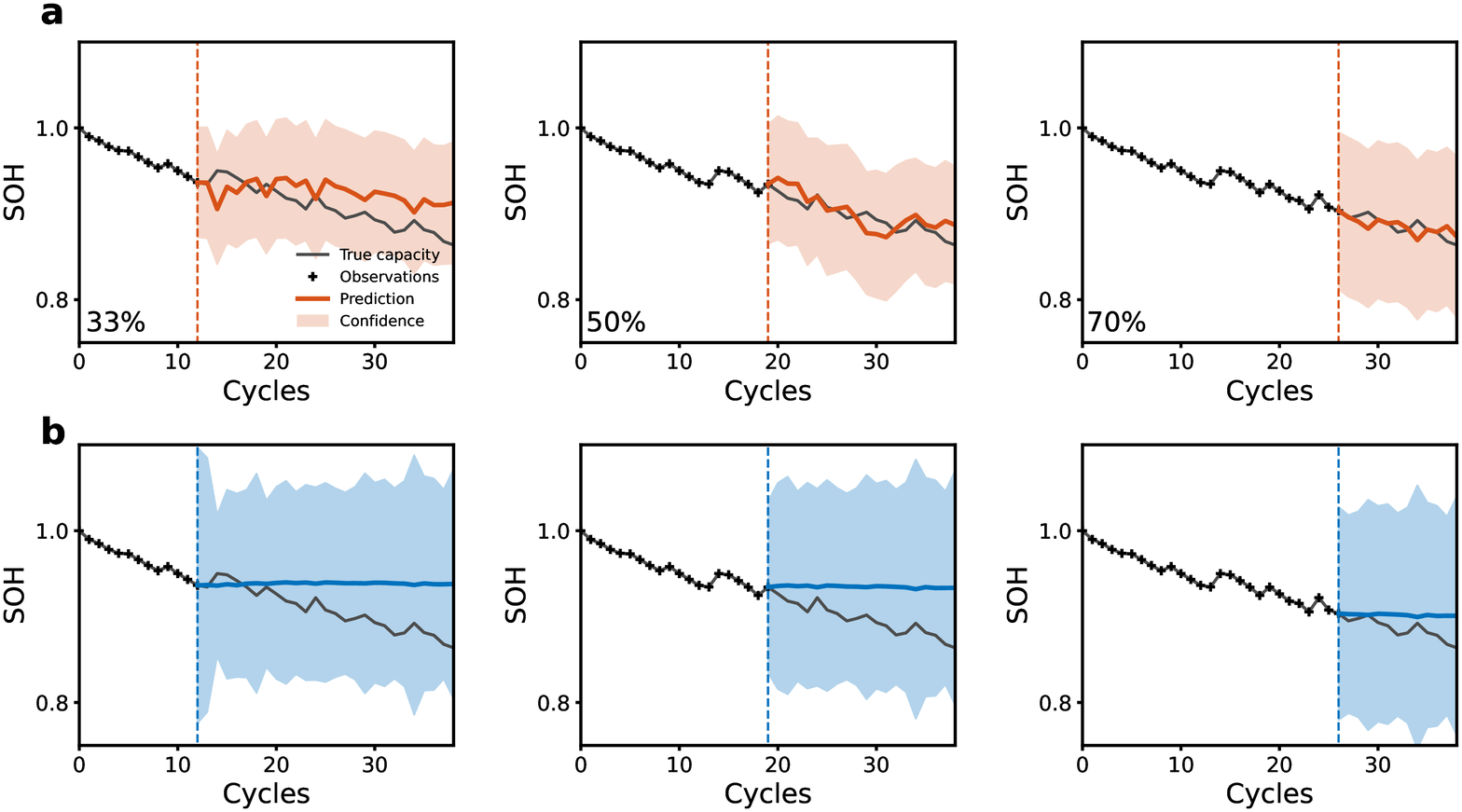}}
	\caption{{State-of-health prediction results for B0032.} (a) GPDM (with transfer learning), (b) GPLVM. The figures from left to right correspond to 33\%, 50\% and 70\% training ratios.}
	\label{Fig10}
\end{figure}

\begin{figure}[h!]
	\centering
	\subfigure
	{\label{fig:c}\includegraphics[width=0.85\linewidth]{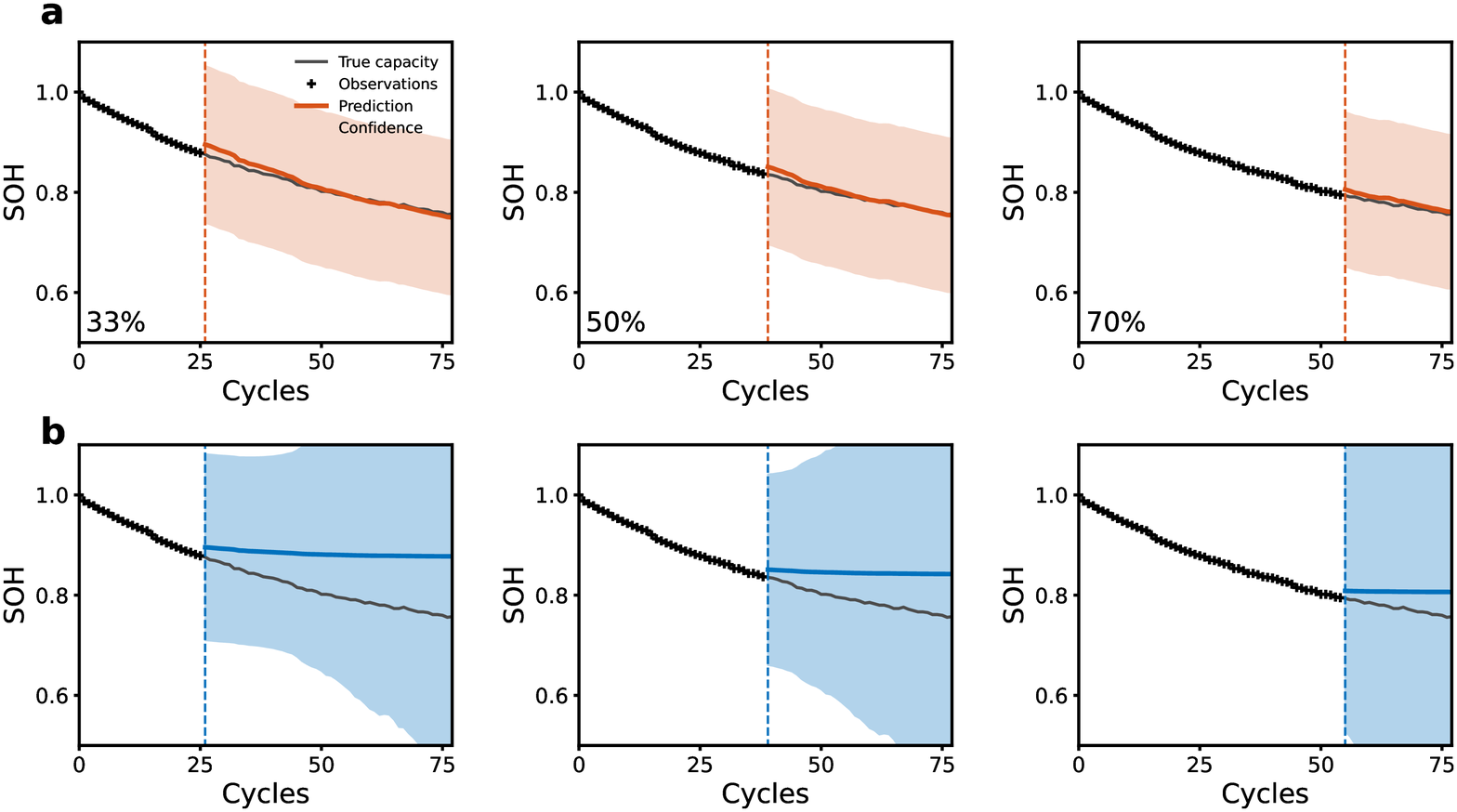}}
	\caption{{State-of-health prediction results for Ox3.} (a) GPDM (with transfer learning), (b) GPLVM. The figures from left to right correspond to 33\%, 50\% and 70\% training ratios.}
	\label{Fig11}
\end{figure}

There are a small number of cases in which the performance noticeably deteriorates. In particular, for battery B0007 we see a significant increase in RMSE for all training point ratios, and likewise for B0032 at 33\%, a seen in Fig. \ref{Fig10}.  Lastly, in the case of Ox3, shown in Fig. \ref{Fig11}, there is a 15\% increase in the RMSE at 30\%.

\begin{figure}[h]
	\centering
	\subfigure
	{\label{fig:c}\includegraphics[width=0.85\linewidth]{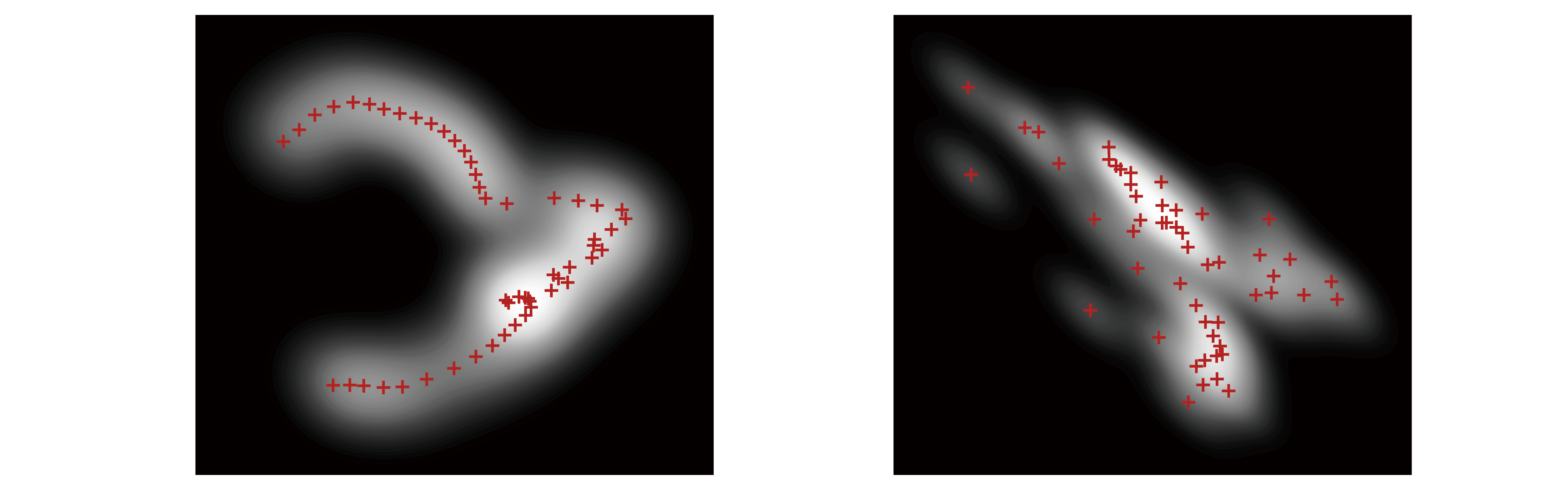}}
	\caption{2-d latent trajectories for B0006 using GPDM (left) and GPLVM (right). The axes correspond to the two latent coordinates and the intensity is proportional to the predictive variance (lighter regions correspond to smaller variance).}
	\label{Fig12}
\end{figure}

One important component of the proposed method is the underlying dynamics as described in Eq. \eqref{eq:linearDM}. If we remove this component, the proposed method is reduced to the  GPLVM. Here, we demonstrate the effects of incorporating the dynamical process by comparing  2-d latent variable reconstructions from  GPLVM and GPDM in the case of B0006 (with attributes as above).  The 2-d latent trajectories $\{\x_n\}$ from the two methods are shown in  Fig.~\ref{Fig12}, with the shading  intensity proportional to the predictive variance (brighter regions indicate smaller variance). 
Fig.~\ref{Fig6} (left) illustrates that GPDM produces a much smoother configuration of latent positions than GPLVM (right), which contains a high degree of scatter. This illustrates that the evolution underlying the state of the battery is much better captured by GPDM, as a consequence of including the dynamics. The latent embeddings provide a good representation of the state of the system.

\section{Conclusions}

Accurate predictions of Li-ion battery stack health status and EOL are crucial for the management and control of these stacks in electric vehicles. One of the issues that has hampered the development of algorithms for early prediction of EOL relates to the incorporation of prior knowledge from other batteries, while also using data that is specific to the battery under consideration. Another issue is in how to incorporate features or attributes, which can vastly improve predictions of the SOH. Here we present a method that overcomes both of these issues in an efficient manner. A  modified GPDM leads to an algorithm with a very low computational burden compared to DNNs, and the added benefit of a confidence interval. 

The results on three data sets show that without additional attributes, GPDM is more accurate in the majority of cases. With the inclusion of additional attributes such as temperature, the accuracy is increased except in a small number of cases. Key to these results is the incorporation of additional data through transfer learning. Without transfer learning it is difficult to capture an overall SOH trend early in the battery life when the trajectory is complex.  

Future research will focus on two aspects: better incorporation of prior knowledge and data for more complex SOH trajectories and the performance of GPDM with more features. The volume and  nature (whether under the same conditions or not) of the data that best provides accurate learning of the SOH trajectory requires a more thorough investigation. Additionally, characterisation of the number and types of features for optimal accuracy requires more extensive data sets combined with rigorous ablation studies.  We suspect that the optimal data and features will be specific to the type of battery. Both of these extensions are beyond the scope of the present study, which serves as a proof of concept for GPDM combined with transfer learning.

\section*{Data statement}
This work involves the use of publicly-available data \cite{saha2007battery,birkl2017oxford}.  Codes will be made available upon request.

\bibliographystyle{unsrtnat2}
\bibliography{mybibfile}

\end{document}